\documentclass[runningheads]{llncs}

\usepackage{eccv}

\usepackage{eccvabbrv}

\usepackage{graphicx}
\usepackage{booktabs}

\usepackage[accsupp]{axessibility}  

\usepackage[pagebackref,breaklinks,colorlinks,citecolor=eccvblue]{hyperref}
\usepackage{hyperref}

\usepackage{orcidlink}

\usepackage[capitalize]{cleveref}
\crefname{section}{Sec.}{Secs.}
\Crefname{section}{Section}{Sections}
\Crefname{table}{Table}{Tables}
\crefname{table}{Tab.}{Tabs.}

\usepackage{graphicx}
\usepackage{amsmath}
\usepackage{amssymb,bm}
\usepackage{booktabs}
\usepackage[normalem]{ulem}

\usepackage[pagebackref,breaklinks,colorlinks]{hyperref}
\usepackage{algpseudocode}
\usepackage{algorithm}

\usepackage[backgroundcolor=blue!20!white,textsize=scriptsize]{todonotes}

\def\Veca{\mathbf{a}}
\def\Vecb{\mathbf{b}}
\def\Vecc{\mathbf{c}}
\def\Vecd{\mathbf{d}}
\def\Vecf{\mathbf{f}}
\def\Vech{\mathbf{h}}
\def\Veck{\mathbf{k}}
\def\Vecn{\mathbf{n}}
\def\Vecp{\mathbf{p}}
\def\Vecq{\mathbf{q}}
\def\Vecr{\mathbf{r}}
\def\Vecs{\mathbf{s}}
\def\Vecu{\mathbf{u}}
\def\Vecv{\mathbf{v}}
\def\Vecw{\mathbf{w}}
\def\Vecx{\mathbf{x}}
\def\Vecy{\mathbf{y}}
\def\Vecz{\mathbf{z}}

\def\Veckappa{\bm{\kappa}}
\def\Vecphi{\bm{\phi}}

\def\VecA{\mathbf{A}}
\def\VecB{\mathbf{B}}
\def\VecC{\mathbf{C}}
\def\VecD{\mathbf{D}}
\def\VecE{\mathbf{E}}
\def\VecF{\mathbf{F}}
\def\VecG{\mathbf{G}}
\def\VecI{\mathbf{I}}
\def\VecJ{\mathbf{J}}
\def\VecK{\mathbf{K}}
\def\VecH{\mathbf{H}}
\def\VecM{\mathbf{M}}
\def\VecP{\mathbf{P}}
\def\VecR{\mathbf{R}}
\def\VecT{\mathbf{T}}
\def\VecU{\mathbf{U}}
\def\VecV{\mathbf{V}}
\def\VecW{\mathbf{W}}
\def\VecX{\mathbf{X}}
\def\VecY{\mathbf{Y}}
\def\VecZ{\mathbf{Z}}

\def\thetaocc{\bm{\theta}_{\rm o}}

\newcommand{\defeq}{\stackrel{\text{\tiny{def}}}{=}}

\DeclareMathOperator*{\argmax}{arg\,max} 
\DeclareMathOperator*{\argmin}{arg\,min} 

\def\transpose{^{\!\mathsf{T}}}
\def\hermitian{^{\!\mathsf{H}}}

\def\gfunc{\textit{g}}
\def\ffunc{\textit{f}}
\def\vfunc{\textit{v}}

\begin{document}

\title{Soft Shadow Diffusion (SSD): Physics-inspired Learning for 3D Computational Periscopy}

\titlerunning{Soft Shadow Diffusion (SSD)}

\author{Fadlullah Raji\orcidlink{0009-0003-0619-0562} \and
John Murray Bruce\orcidlink{0000-0002-1416-4175}}

\authorrunning{Raji and Murray-Bruce}

\institute{ Computer Science and Engineering, University of South Florida\\ 
4202 E. Fowler Avenue, Tampa, FL, USA 33620.\\
{\tt\small \{fraji, murraybruce\}@usf.edu }
{\tt\small } \\}

\maketitle

\begin{abstract}
Conventional imaging requires a line of sight to create accurate visual representations of a scene.
In certain circumstances, however, obtaining a suitable line of sight may be impractical, dangerous, or even impossible.
Non-line-of-sight (NLOS) imaging addresses this challenge by reconstructing the scene from indirect measurements.
Recently, passive NLOS methods that use an ordinary photograph of the subtle shadow cast onto a visible wall by the hidden scene have gained interest. These methods are currently limited to 1D or low-resolution 2D color imaging or to localizing a hidden object whose shape is approximately known. Here, we generalize this class of methods and demonstrate a 3D reconstruction of a hidden scene from an ordinary NLOS photograph. To achieve this, we propose a novel reformulation of the light transport model that conveniently decomposes the hidden scene into
\textit{light-occluding} and \textit{non-light-occluding} components to yield a separable non-linear least squares (SNLLS) inverse problem. We develop two solutions: A gradient-based optimization method and a physics-inspired neural network approach, which we call Soft Shadow diffusion (SSD).
Despite the challenging ill-conditioned inverse problem encountered here, our approaches are effective on numerous 3D scenes in real experimental scenarios. Moreover, SSD is trained in simulation but generalizes well to unseen classes in simulation and real-world NLOS scenes. SSD also shows surprising robustness to noise and ambient illumination.
\end{abstract}
\begin{figure}[ht!]
    \centering
    \includegraphics[width=0.99\linewidth]{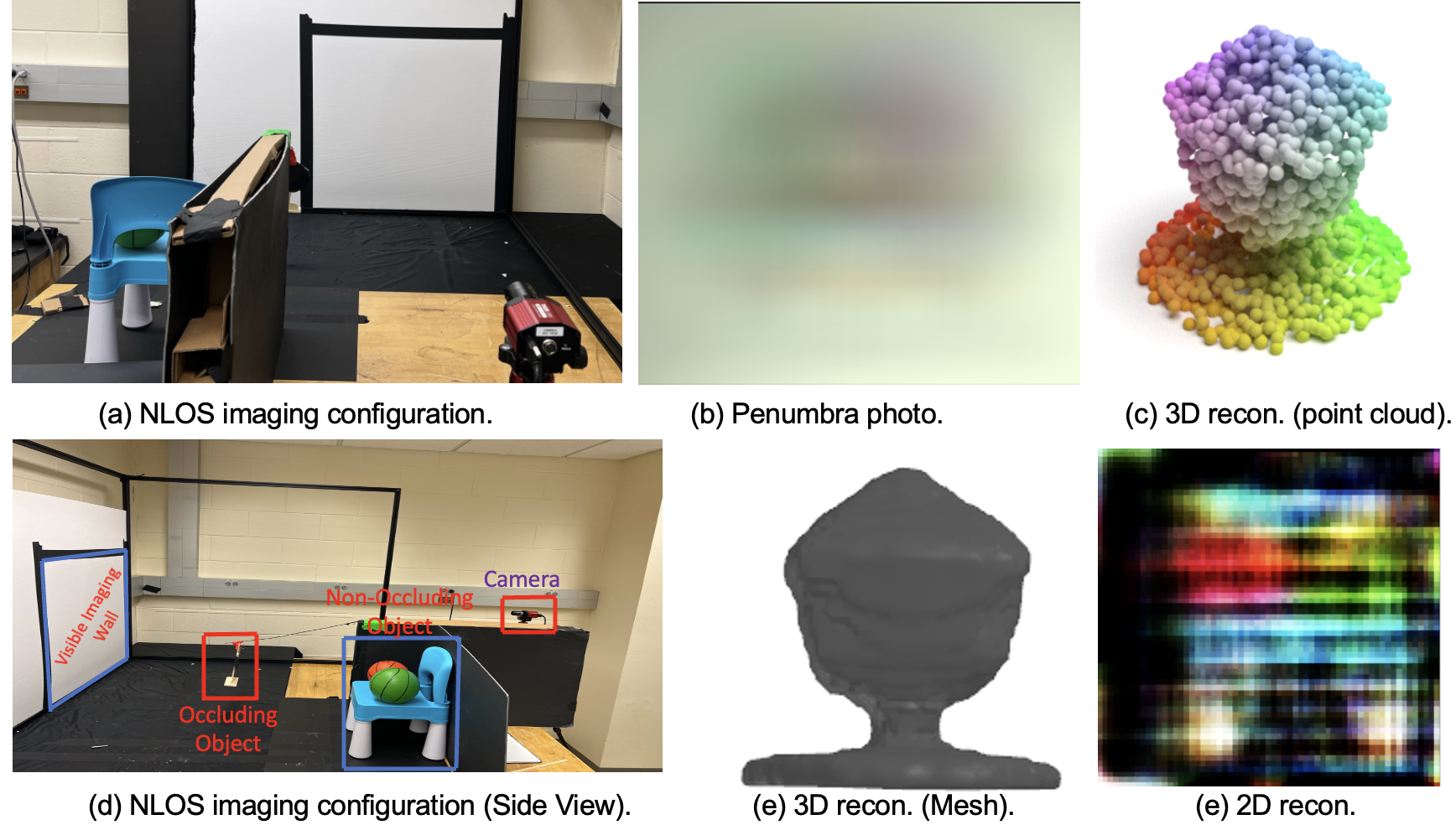}
    \caption{\textbf{NLOS imaging configuration.}
    Non-occluding objects diffusely reflect light toward the visible wall; the reflected light is partially occluded by an occluding object in the hidden scene to create a soft shadow on the visible wall.%
    }
    \label{fig:nlos_config}
\vspace{-5mm}
\end{figure}
\vspace{-5mm}
\section{Introduction}
Collision avoidance systems in autonomous navigation, detection of hidden adversaries in military operations, danger assessment in search-and-rescue operations, and arterial blockage assessment in biomedical imaging are examples of applications that will benefit from the ability to form images of scenes beyond an observer’s line of sight.
Because of its breadth of potential applications, this field called non-line-of-sight (NLOS) imaging continues its rapid growth. The fundamental underlying goal is to infer information about the NLOS scene from measurements of signals, such as light~\cite{Kirmani2012, Faccio2020non,saunders2019computational,Saunders2020multi}, sound~\cite{Lindellaccoustic,Boger2023towards}, heat~\cite{maeda2019thermal, ThermalNLOS}, or other modality~\cite{Adib2013see}, that reach the observer after being diffusely reflected or emitted from the NLOS scene.

In this work, we demonstrate 3D NLOS imaging from measurements of the light reaching a surface that is visible to the observer and the hidden scene.
Prior methods that similarly rely on measurements of light fall into either of two categories:
\textit{active methods}, which exploit controlled illumination of the hidden scene~\cite{Kirmani2012,velten2012,pawlikowska2017single,Otoole2018confocal,Heide2019OcclusionNLOS,lindell2019wave,liu2019non,rapp2020edge-resolved,liu2020phasor,Metzler2020deep,liu2022fewshot,Faccio2020non,Seidel2023non,somasundaram2023role}; or
\textit{passive methods}, which exploit already existing light without any control over the hidden scene illumination~\cite{Bouman2017,Baradad2018inferring,saunders2019computational,murray2019occlusion,Yedidia_2019_CVPR,seidel2019corner,Seidel2021_two,Geng2022passive,Wang2021accurate,Czajkowski2024}.
Active methods have demonstrated remarkable success in achieving high-resolution 3D NLOS imaging \cite{Otoole2018confocal, velten2012, pawlikowska2017single}, by using ultrafast pulsed-laser illumination, and time-resolved single-photon detection.
Despite these advances, the complexity and cost of the equipment required by active methods can be a limiting factor
in several applications.
In contrast, passive methods offer a simpler alternative but are currently limited to low-resolution 1D~\cite{Bouman2017,seidel2019corner} or 2D~\cite{saunders2019computational, Saunders2020multi, Yedidia_2019_CVPR, Seidel2021_two} reconstructions of hidden scenes.
A recent method \cite{Czajkowski2024,Czajkowski2024_teri3D} exploits two
orthogonal
edges of a door frame to achieve remarkable 3D imaging of non-occluding, hidden-scene objects. However, the exploited shadows are cast by the visible, known door frame occluder.
Here, \textbf{we generalize occluder-aided passive methods to include 3D imaging of occluders that are hidden or otherwise unknown}.

\noindent
\textbf{Contributions: }In this paper, we propose a reformulation of the computational periscopy as a separable non-linear least squares (SNLLS) inverse problem that enables 3D reconstructions from a single penumbra photograph, and we offer a gradient-based optimization to solve it.
In addition, we introduce \textit{Soft Shadow Diffusion} (SSD) a novel generative model based on denoising probabilistic diffusion for reconstructing 3D shapes, as high-resolution 3D point clouds, from a single photograph of their soft shadows.
Finally, we experimentally demonstrate the first joint reconstructions of the 3D occluding and the 2D non-occluding objects in a real experimental scene.

\subsection{Related work}
\subsubsection{Computational Periscopy:}
\Cref{fig:nlos_config}(a) shows a photograph of the NLOS imaging scenario considered here and by prior related works (e.g.,~\cite{Baradad2018inferring,saunders2019computational,Yedidia_2019_CVPR,Saunders2020multi}).
In this scenario, a preexisting ambient light source
illuminates the hidden scene area (and the \textit{visible wall} in the observer's LOS).
Light reflected, toward the visible wall, from some parts of the hidden scene
(here, the chair and basketballs) is partially occluded by other hidden scene objects (here, the spade-like shape) producing a penumbra (or soft shadow) on the visible wall.

From simple photographs of penumbra, Saunders et al.~\cite{saunders2019computational,Saunders2020multi} and Baradad et al.~\cite{Baradad2018inferring}
produce reconstructions of the light-reflecting (non-light-occluding) portions of the hidden scene by assuming that the light-occluding hidden scene structures have a known shape but unknown location, or that projections of the occluder on the visible wall are available from a prior calibration step. Furthermore, by assuming that the occluder is approximately planar,
Yedidia et al.~\cite{Yedidia_2019_CVPR}
reconstruct its 2D projection and a 2D video of the non-light-occluding portion of the hidden scene from a video of the visible penumbra.
In this paper, we develop a new computational strategy to reconstruct the 3D structure of the light-occluding objects from a 2D photograph of penumbra.
Hence, the output computed by our method can augment these prior approaches by providing the necessary prior knowledge about the hidden occluders.

Instead of reconstructing its image, directly inferring the contextual properties of a hidden scene may be important in various applications. Along those lines, Sharma et al. proposed learning-based methods for accurately determining human activity or classifying the number of people in an unknown room from a video recording of penumbra~\cite{Sharma2021what}. Medin et al. further showed that a single penumbra encodes enough information for biometric identification of an individual without a line of sight~\cite{Medin2023can}.
Although direct photographs of sharp human silhouette shadows have long been explored for biometrics~\cite{Iwashita2012gait,Verlekar2018gait}, Medin et al.~\cite{Medin2023can} consider a challenging variation: The observed shadow is instead of an object that occludes light reflected by the face of the hidden individual.
Rather impressively, both methods~\cite{Sharma2021what,Medin2023can} demonstrate accurate inference without prior knowledge of the hidden occluder or scene geometry.
\vspace{-3mm}
\subsubsection{Shape from S:}
Shape from shading \cite{hornshapesharding, Ramachandran1988PerceptionOS} and shape-from-shadow \cite{cavanagh1989shape} are areas of longstanding and broad interest to the computer vision community.
Shape from shading reconstructs a 3D shape from its 2D image by analyzing variations in brightness (shading) across the image, to deduce the orientation of each point on the object's surface.
In contrast, shape from shadow reconstructs a 3D shape from several (binary) shadow images, created by varying the illumination direction, by analyzing the darkest parts across the image stacks to deduce the depth and surface orientations.
Recent approaches have exploited data-driven approaches to yield impressive reconstructions \cite{deepshadow, Yamashita2010, Jingwang2023 }, even in a monocular setting \cite{Chaudhury2024ShapeShading}.
These prior works, exploit cues from shading due to direct illumination and sharp shadows caused by mutual occlusion under known or possibly unknown illuminations. Here, we use a single image of the soft shadow created on a nearby surface, by hidden 3D occluding objects partially blocking a complex, diffuse, and hidden illumination both lying outside the image's field of view.
\vspace{-3mm}
\subsubsection{Diffusion Probabilistic Models:}

Generative diffusion probabilistic models \cite{diffusionfoundation, song2019generative, DDPM}. These models have demonstrated high-quality image generation \cite{Rombach_latent_2022_CVPR, Yang_2023_CVPR}, rivaling generative adversarial networks (GANs)\cite{dhariwal2021diffusion}. Their application has broadened to include tasks such as time-series forecasting \cite{timeseriesdiffusion}, audio generation \cite{diffwave}, text-to-speech \cite{wavediff2, text2speech}.
Several works have explored these models for point cloud reconstructions via a two-stage generation using flow models~\cite{yang2019} or diffusion priors~\cite{cai2020, zeng2022}. To reduce generation latency, Zhou et al.~\cite{zhou20213d} proposed PVD, a single-stage diffusion prior for generating point clouds. Beyond point clouds, text-conditional 3D generation has advanced with methods like DreamFields~\cite{jain2021} and CLIP-guided mesh optimization~\cite{Khalid2022}. Other approaches focus on reconstructing 3D from multi-view images using diffusion-based priors~\cite{ poole2022dreamfusion, liu2023zero1to3, ye2024consistent1to3}. 

This work is a novel extension of diffusion models to generate 3D point clouds from a 2D soft shadow photograph. We subsequently use a trained regression model to generate a 3D mesh representation of the object casting the shadow. Building on prior work that performs diffusion over individual points, our approach conditions the diffusion process on soft shadow images.

\section{Light Propagation Model}
\label{sec:overview}
Consider the configuration shown in \Cref{fig:nlos_config}(a)
and let $\Vecx$ denote any point in the hidden scene volume.
A hidden scene point whose light paths to the visible wall portion within the camera's FOV are (un)occluded along some directions, but not all, is well-conditioned for recovery.
The set $\cal X$ of all such hidden scene points forms the \textit{computational FOV} (CFOV) of the imaging system~\cite{saunders2019computational}.
Let $f(\Vecx)$ denote the unknown radiosity distribution at a hidden scene point $\Vecx$, then the irradiance $i(\Vecp)$ of a visible wall point $\Vecp$ is\cite{saunders2019computational}:
\begin{align}
i(\Vecp) = \int_{\Vecx \in \mathcal{X}} \frac{g(\Vecp, \Vecx)}{\| \Vecx - \Vecp\|^2_2} v(\Vecx,\Vecp;\bm{\theta}_{\mathrm{o}}) f(\Vecx)\, \mathrm{d}\Vecx + b(\Vecp),
\label{eqn:forwardmodel}
\end{align}
where $ g(\Vecp, \Vecx) = \cos(\measuredangle(\Vecx -\Vecp, \Vecn_{\Vecx})) \cos(\measuredangle(\Vecp- \Vecx, \Vecn_\Vecp)) $ models
Lambertian reflection and foreshortening with
$ \Vecn_\Vecp $ and $ \Vecn_\Vecx $
representing the surface normals at $\Vecp$ and $\Vecx$,
respectively.
The function $v(\Vecx, \Vecp; \bm{\theta}_{\rm o})$ models the visibility between $ \Vecp $ and $ \Vecx $ for an unknown occluding object whose shape is implicitly parameterized by $\bm{\theta}_{\rm o}$; $v(\Vecx, \Vecp; \bm{\theta}_{\rm o}) = 1$ when the ray from $\Vecx$ to $\Vecp$ is unoccluded, otherwise it is $0$.
Finally, $b(\Vecp)$ models contributions due to noise,
and other sources of illumination---predominantly any existing visible side illumination sources and illumination from hidden scene points outside the CFOV, i.e.$\int_{\Vecx \notin \mathcal{X}} \frac{g(\Vecp, \Vecx)}{\| \Vecx - \Vecp\|^2_2} f(\Vecx)\, \mathrm{d}\Vecx$---that do not form penumbra.
The recoverable hidden scene $\mathcal{X}$ is typically assumed to be a single plane at a known depth~\cite{saunders2019computational,Aittala2019computational,Yedidia_2019_CVPR}, or a few planes~\cite{Baradad2018inferring,Saunders2020multi} at known depths, from the visible wall.
Discretizing the measurement plane and the hidden scene plane(s) into pixels gives the discrete forward model: 
\begin{align}
   \Vecy = \VecA(\bm{\theta}_{\rm o})\Vecf + \Vecb,
\label{eq:discrete_forward_model}
\end{align} 
where $\Vecy \in \mathbb{R}^M$ is a vectorization of the measured photograph, $\Vecf \in \mathbb{R}^N$ is the vector containing the unknown radiosities of the hidden scene pixels, and $\VecA(\bm{\theta}_{\rm o})\in\mathbb{R}^{M\times N}$ is the light transport matrix given $\bm{\theta}_{\rm o}$ (which implicitly represents the shape and position of the occluding objects).

Saunders et al.~\cite{saunders2019computational} assume a known occluder shape but use $\bm{\theta}_{\rm o}$ to represent its location.
This work considers the case where the occluder's shape is also unknown and uses $\bm{\theta}_{\rm o}$ as an implicit representation for the 3D shape of an unknown occluder.
Estimating the unknowns $\bm{\theta}_{\rm o}$, $\Vecf$ and $\Vecb$ using \eqref{eq:discrete_forward_model} is an ill-conditioned \textit{separable nonlinear inverse problem}.
In \Cref{sec:inversion} we present two solutions for this inverse problem by building on our model and explicitly delineating the hidden scene as comprising two components: those that act as occluders of light, and those that act only as emitters/reflectors of light.

\begin{figure*}[ht!]
    \centering
  \begin{subfigure}{.32\textwidth}
    \centering
    \includegraphics[width=1\linewidth]{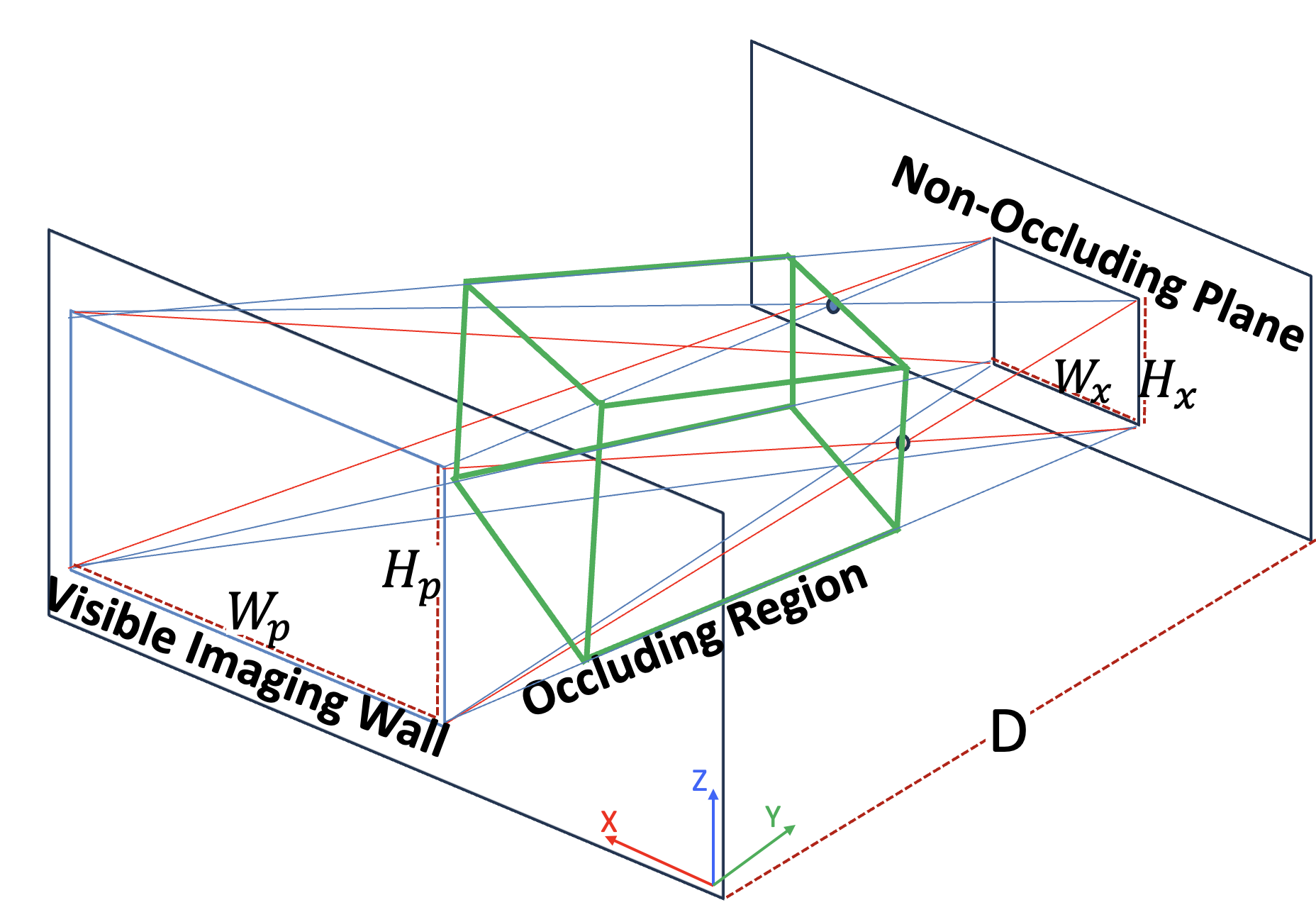}
    \caption{Computational FOV}
    \label{fig:computational_fov}
  \end{subfigure}
  \hfill
  \begin{subfigure}{.32\textwidth}
    \centering
    \includegraphics[width=1\linewidth]{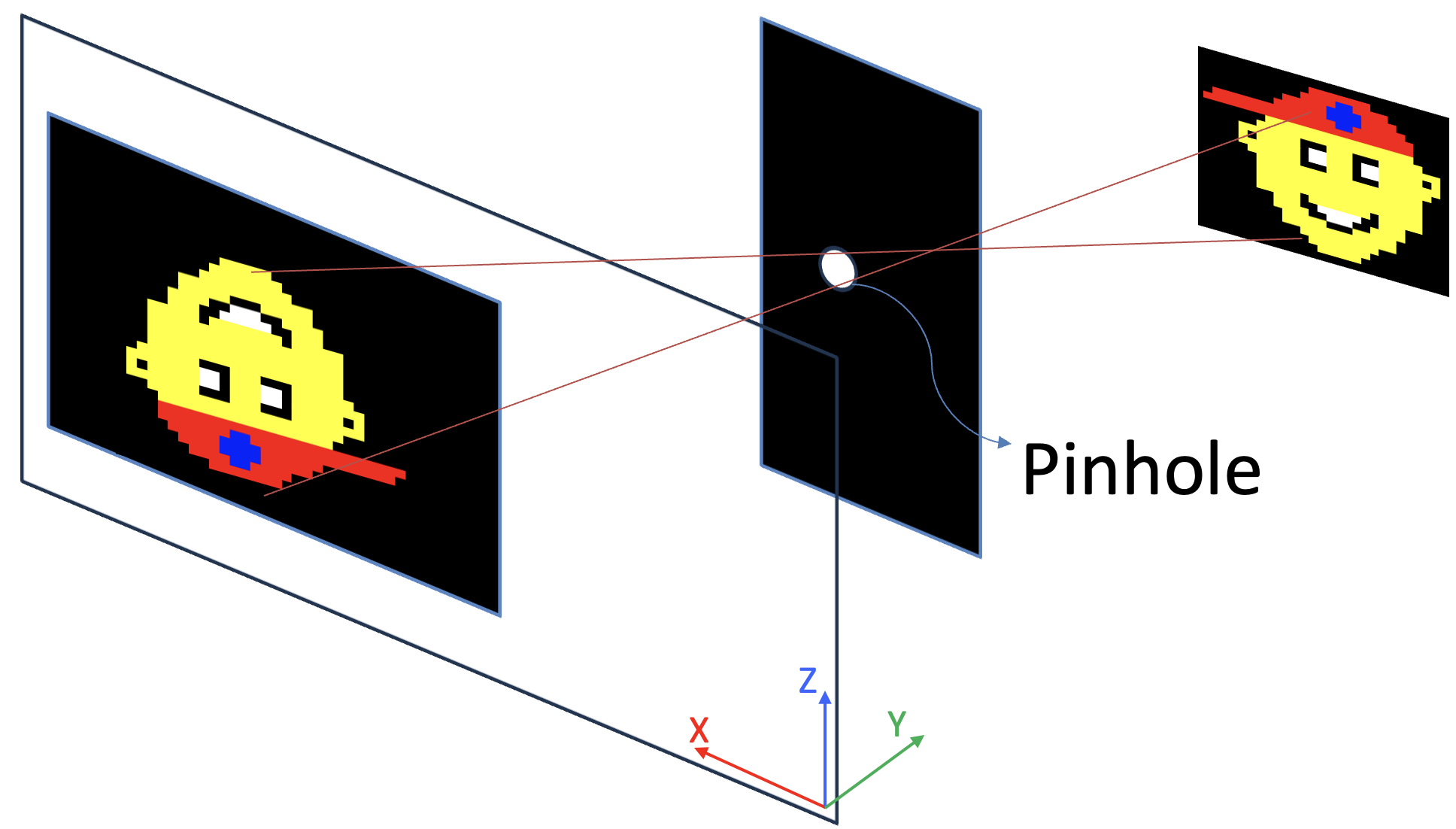}
    \caption{A pinhole occluder}
    \label{fig:pinhole}
  \end{subfigure}
  \hfill
  \begin{subfigure}{.32\textwidth}
    \centering
    \includegraphics[width=1\linewidth]{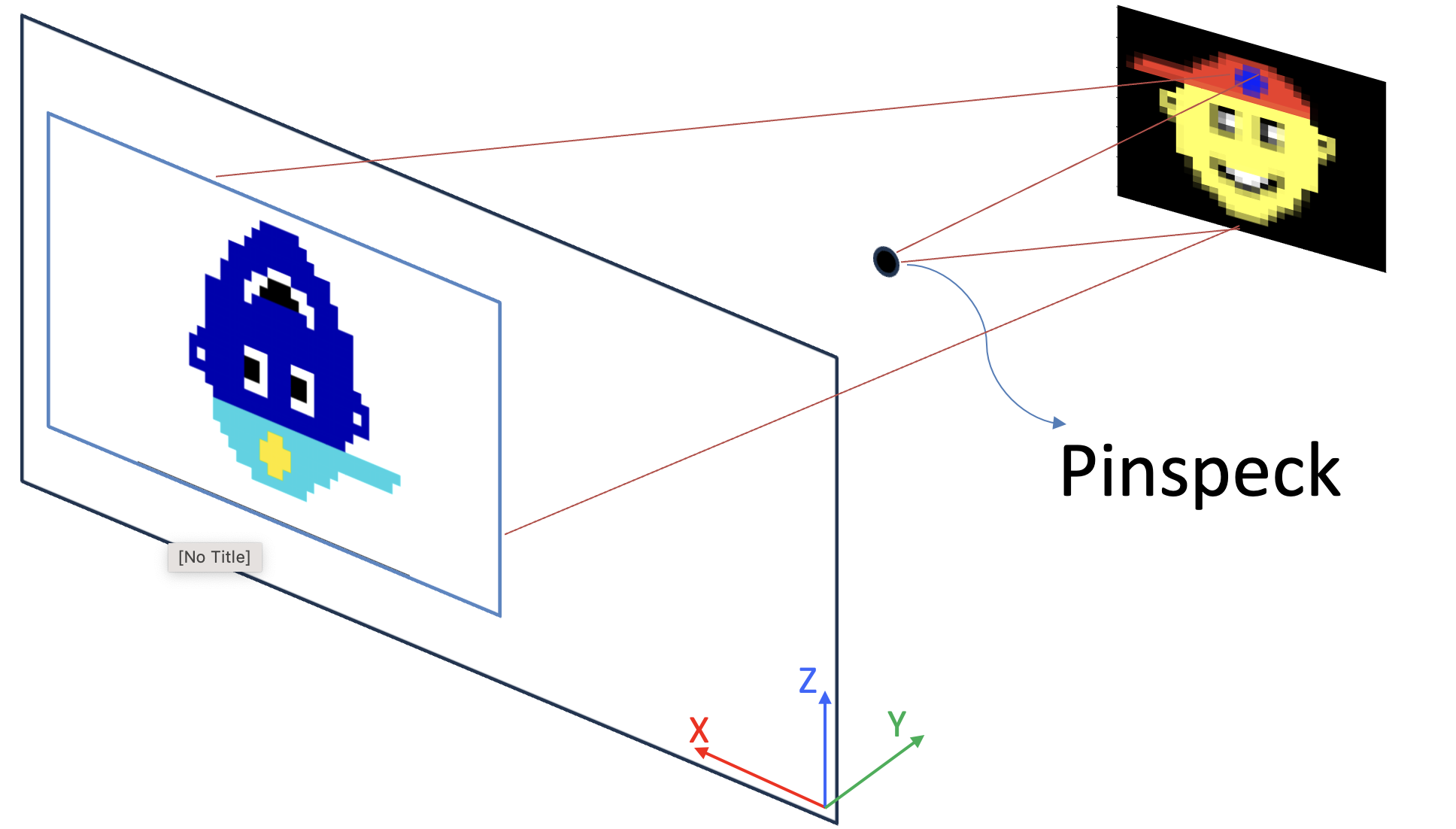}
    \caption{A pinspeck occluder}
    \label{fig:pinspeck}
  \end{subfigure}
    \caption{\textbf{Computational FOV and two simple examples of occluders.} (a) The camera's FOV is projected towards the non-occluding scene plane.
    The occluding region is the volume created by joining the boundaries of the measurement FOV to the boundaries of the desired hidden scene plane.
    (b) An ideal pinhole projects an image of the scene. (c) An ideal pinspeck projects a negative image of the scene.}
    \label{fig:cfov_occl}
\vspace{-8mm}
\end{figure*}
%
%
%
%
\subsection{Hidden Scene Decomposition and Computational FOV}
\label{ssec:cfov}
We aim to reconstruct simultaneously a 3D image of the hidden occluders and a 2D image of the hidden non-occluders (i.e., emitters/reflectors of light).
To obtain a feasible inverse problem and facilitate accurate reconstructions, we must restrict the reconstruction volume to include only hidden scene portions that can cast penumbra somewhere within the photographed portion of the visible wall.
The 2D image of the non-occluders is assumed to be the radiosity of a plane of fixed size $W_{\rm x} \times H_{\rm x}$ at a depth $D$ from the visible wall (see \Cref{fig:computational_fov}).
Then, as illustrated in \Cref{fig:computational_fov}, the \textit{occluding volume} is then defined by the frustum whose edges coincide with the lines formed by connecting each vertex of the camera FOV to each vertex of the non-occluding hidden scene plane.

For our problem setting, the 3D \textit{occluding volume} and the chosen 2D \textit{non-occluding plane}, are collectively referred to as the computational FOV (CFOV).
This represents a natural extension of the CFOV concept~\cite{saunders2019computational} from 2D to 3D.

\section{3D Computational Periscopy: Models \& Representation}
\label{sec:methods}

To facilitate 3D reconstruction using our new reformulation, we consider two different representations of 3D structures. The first, presented in \Cref{ssec:voxel_uniform}, is an explicit representation that uses a uniform voxelization
of the occluding frustum portion of the CFOV. The second, presented in \Cref{ssec:ssd}, is an implicit neural point cloud-based representation, which imposes a \textit{deep prior} for 3D shapes.
Each representation inspires a corresponding reconstruction method presented in \Cref{sec:inversion} and builds upon a simple pinspeck occluder.

\subsection{A Simple Pinspeck Occluder}
\label{ssec:pinspeck_model}
A pinspeck (or anti-pinhole) occluder~\cite{Cohen1982,Torralba2014}
is the optical complement of a pinhole. Thus, a single \textit{pinspeck} at $\Vecu = (u_x, u_y, u_z)$
blocks all rays that reach the point $\Vecu$
 such that
the visibility between a point $\Vecp = (p_x, 0, p_z)$ on the visible wall, and a point $\Vecx = (x, y, z)$ on the non-occluding hidden scene plane is
\begin{align}
    v_{\text{ps}}(\Vecp,\Vecx;\Vecu) =
    &1 {-} \delta\!\left(p_x {-} m u_x {-} \frac{xy}{u_y} \right) \delta\!\left( p_z {-} m u_z {-} \frac{yz}{u_y}  \!\right)
    \label{eq:revised_pinspeck_visibility}
\end{align}
where $\delta(\cdot)$ is the Dirac delta function, and $m = 1 - y/u_y $. Imposing the optical complementarity property~\cite{Cohen1982,Torralba2014}, gives the following expression for the (negative) image formed on the visible plane:
\begin{align}
    i(\Vecp) 
    &= \int_{\Vecx \in \mathcal{X}} \frac{g(\Vecp, \Vecx)}{\| \Vecx - \Vecp\|^2_2} \left( 1 - v_{\text{ph}}(\Vecp,\Vecx;\Vecu)\right) \, f(\Vecx)\, \mathrm{d}\Vecx,
    \label{eq:revised_pinspeck_model}
\end{align}
where the pinhole visibility $v_{\text{ph}}(\Vecp,\Vecx;\Vecu) = 1 - v_{\text{ps}}(\Vecp,\Vecx;\Vecu)$.

\subsection{Uniform Voxels Representation}
\label{ssec:voxel_uniform}
Following our prior work~\cite{Raji2024towards}, assume the light-occluding frustum is uniformly discretized into $K$ voxels whose centers are located at $\{\Vecu_k\}_{k=1}^K$, then with a dense discretization, arbitrary 3D shapes can be well-approximated by a subselection of voxels identified by their indices.
In this scenario, $ v(\Vecx,\Vecp;\thetaocc) = 1 -\sum_{k=1}^{K} \alpha_k v_{\text{ph}}(\Vecp,\Vecx;\Vecu_k) $ where the binary-valued scalar $\alpha_k$ indicates the presence (1) or absence (0) of a pinspeck at $\Vecu_k$, such that the visible wall irradiance is approximately:
\begin{align}
    i(\Vecp)  &= \!\!\int_{\Vecx \in \mathcal{X}}\!\! \frac{g(\Vecp, \Vecx)}{\| \Vecx {-} \Vecp\|^2_2}\!\! \left(\!1 {-}\!\! \sum_{k=1}^{K} \alpha_k v_{\text{ph}}(\Vecp,\Vecx;\Vecu_k)\!\right)\!\! f(\Vecx)\, \mathrm{d}\Vecx.
    \label{eq:revised_multipinspeck_model}
\end{align}

We discretize the non-occluding plane into $N = (N_{\rm x} \times N_{\rm z})$ pixels and represent the unknown radiosity of the $n$-th pixel by $f_n$ and the collection of all $N$ radiosities by $\Vecf \in \mathbb{R}^N$. Additionally, we define $ \{\Vecp_m\}_{m=1}^M $ as the locations of pixels on the visible wall that are captured by the $ M = (M_{\rm x} \times M_{\rm z}) $ camera pixels. Thus,
\begin{equation}
    \Vecy = \VecA\!\!\left(\!1{-}\sum_{k=1}^K\alpha_k \VecV_k\!\right)\!\Vecf = \left(\!\VecA {-} \sum_{k=1}^K\alpha_{k}\VecA\VecV_k\right)\Vecf,
    \label{eq:new_discrete_forward_model}
\end{equation}
where $\Vecy \in \mathbb{R}^M $ is the vectorized penumbra photograph, $ \VecV_k \in \mathbb{R}^{M\times N} $ is the matrix corresponding to visibility of a pinhole at $\Vecu_k$.
Reconstructing the 3D shape is, thus, equivalent to computing $ \thetaocc\defeq\bm{\alpha} =  (\alpha_1, \alpha_2,\ldots, \alpha_K){\transpose}$ from $\Vecy$.

\subsection{Point Cloud Representation}
\label{ssec:pc_representation}
While a uniform voxelization has the potential to provide accurate scene representations with finer discretizations; it, however, leads to increased computational complexity, memory requirements, and poor conditioning. Furthermore, a binary constraint on $\bm{\alpha}$ yields a nonconvex optimization. In contrast, a point cloud representation deviates from the uniform grid-based, directly optimizing the vector locations $\Vecu_k$ instead of the binary vector $\bm{\alpha}$. It also provides a lower-dimensional representation that effectively captures a wide range of 3D shapes while being compatible with our pinspeck-based model~\eqref{eq:revised_pinspeck_model}. Thus, we aim to solve $\Vecy = \VecA(\thetaocc)\Vecf$, where $\thetaocc\defeq\{\Vecu_k\}_{k=1}^K$.
Directly optimizing the locations of $K$ points is challenging. Nonetheless, to facilitate accurate and high-resolution reconstructions, we develop and exploit a learned point cloud shape prior, which we call \textbf{\textit{soft shadow diffusion}}.
\subsection{Soft Shadow Diffusion Model}
\label{ssec:ssd}
\begin{figure}[t!]
    \centering
  \begin{subfigure}{.5\textwidth}
    \centering
    \includegraphics[width=1\linewidth]{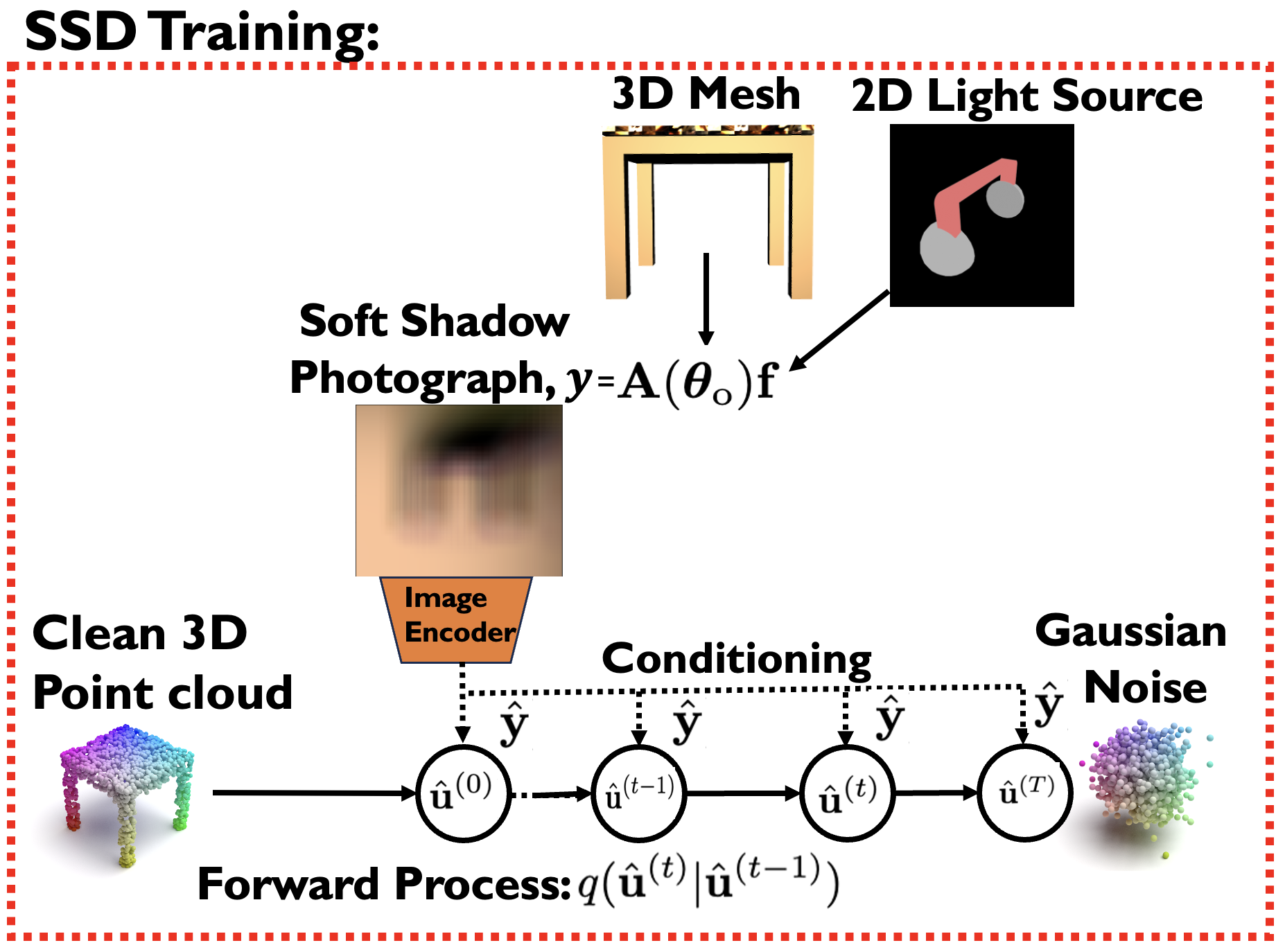}
    \caption{Training Pipeline}
    \label{fig:ssd_training}
  \end{subfigure}%
  \begin{subfigure}{.5\textwidth}
    \centering
    \includegraphics[width=1\linewidth]{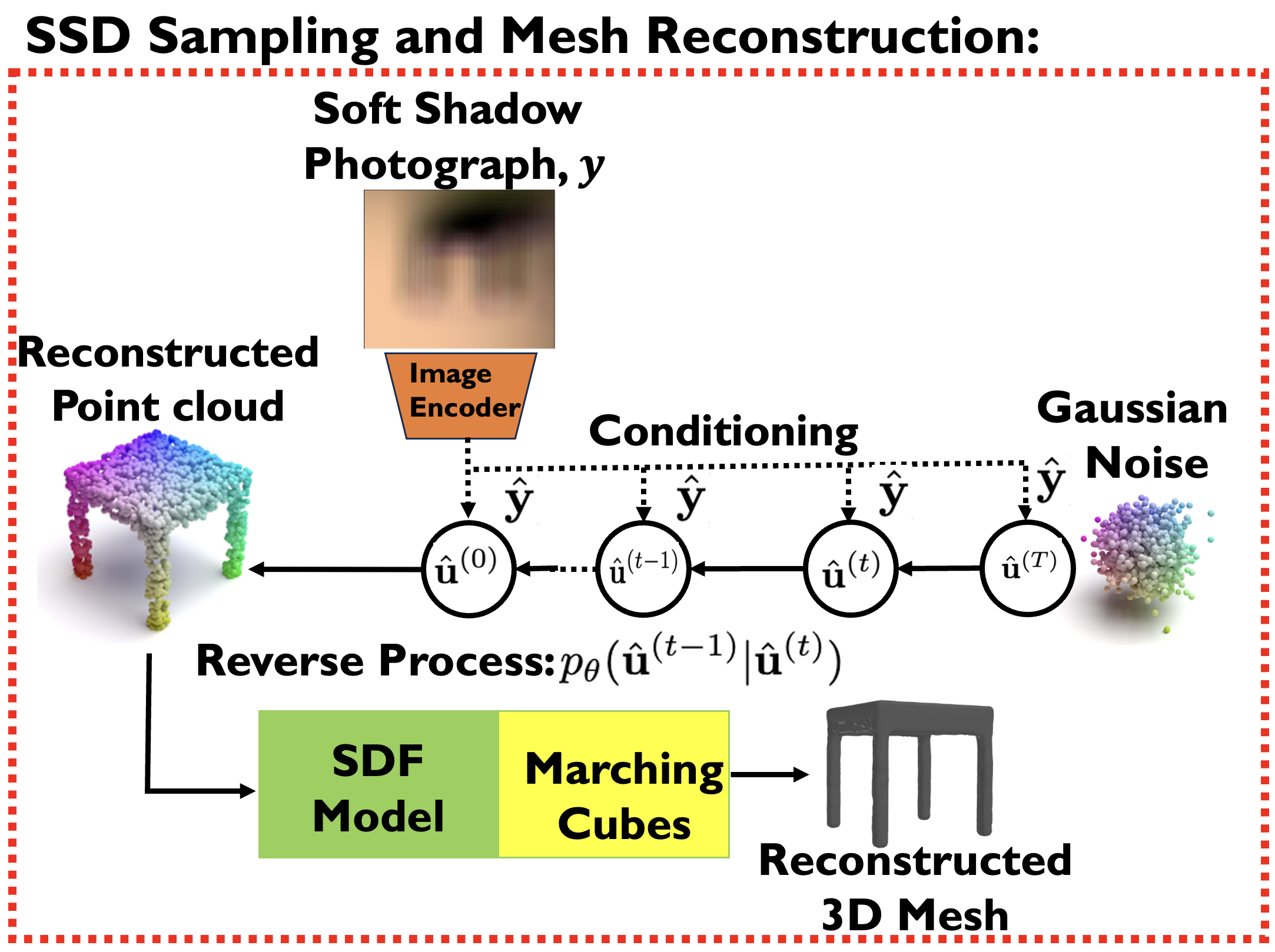}
    \caption{Inference Pipeline}
    \label{fig:ssd_sampling}
  \end{subfigure}
    \caption{\textbf{Soft Shadow Diffusion Model Pipeline.}
    (a) In training, the model maps a clean input point cloud to noise using the forward diffusion process. (b) At inference, the model starts from an initial noise and transforms this noise into a point cloud representation of light-occluding structure by conditioning on the encoded soft shadow photograph $\hat{\Vecy}$. The point cloud is then converted into a mesh representation of the occluding structure.
    }
    \label{fig:ssd_model}
\vspace{-5mm}
\end{figure}
Our proposed soft shadow diffusion (SSD) model adapts the state-of-the-art denoising diffusion models \cite{diffusionfoundation, luo2021diffusion, zeng2022lion, Shim_2023_CVPR}
to conditional point cloud generation. Here, the denoising diffusion process is conditioned on a latent representation $\hat{\Vecy}$
of a measured soft shadow photograph $\Vecy$.
Our motivation is to leverage SSD as a powerful learned prior for high-resolution 3D point clouds that explain the soft shadows in a penumbra photograph. 

Once trained, our SSD model $G_{w}(\Vecy)$ is used to generate a denoised point cloud $ \{ \Vecu_k^0\}_{k=1}^K$ by iteratively reversing a Markov forward process, as depicted in \Cref{fig:ssd_sampling}.
Specifically, conditioned on a soft shadow photograph latent $\hat{\Vecy}$,
the model iteratively reduces the noise from an initial point cloud $\{\Vecu_k^T\}_{k=1}^K$ that is a random realization of a Gaussian noise process.

The training process (see \Cref{fig:ssd_training}) begins by setting $ \{\Vecu_k^{0}\}_{k=1}^K$ to be a random sample from our dataset of noise-free point clouds.
Then, a series of progressively noisier point clouds $ \{\Vecu_k^{t}\}_{k=1}^K$ for $t = 1, 2, \ldots, T$ are computed from $\{\Vecu_k^{0}\}_{k=1}^K$ using $ \Vecu_k^{t} = \sqrt{\bar{\alpha}_t}\Vecu_k^{0} + \sqrt{1 - \bar{\alpha}_t}\epsilon $ for $k=1,2,\ldots, K$, where $ \epsilon $ is zero-mean white Gaussian noise with unit variance, and the coefficient $ \alpha_t $ gradually decreases as the time step $ t $ increments. With these noisy samples, the weights $w$ of the underlying neural network are updated by minimizing the following loss function \cite{song2019generative}: 
\[
L = \mathbb{E}_{t,\{\Vecu^0_k\}_{k=1}^K,\epsilon} \left\| \epsilon_\theta(\{\Vecu_k^t\}_{k=1}^K, \hat{\Vecy}, t) - \epsilon \right\|^2_2.
\label{eq:loss_function}
\]

We provide additional details on training and sampling from our SSD model in the supplement.

\section{3D Computational Periscopy: Inversion}
\label{sec:inversion}
Assuming background contributions to the term $\Vecb$ is small,
a solution to \eqref{eq:discrete_forward_model} can be obtained by solving the Tikhonov-regularized minimization
\begin{equation}
    \argmin_{(\Vecf, \thetaocc)} \left \| \VecA(\bm{\theta}_{\rm o})\Vecf  - \Vecy\right\|_2^2 + \lambda\|\Vecf\|_2^2,
    \label{eq:tik_reg}
\end{equation}
where $\| \cdot \|_2$ is the 2-norm, and $\lambda>0$ is the regularization parameter.
An alternating approach for the minimization \eqref{eq:tik_reg} is,
\begin{align}
    \widehat{\Vecf}^k &= \argmin_{\Vecf} \left \| \VecA(\widehat{\bm{\theta}}_{\rm o}^{k-1})\Vecf  - \Vecy\right\|_2^2 + \lambda\|\Vecf\|_2^2\nonumber\\
                        &
                        = \left(\VecA{\transpose}(\widehat{\bm{\theta}}_{\rm o}^{k-1})\VecA(\widehat{\bm{\theta}}_{\rm o}^{k-1}) + \lambda \VecI \right)^{-1}\VecA{\transpose}(\widehat{\bm{\theta}}_{\rm o}^{k-1})\Vecy\\
    \widehat{\bm{\theta}}_{\rm o}^{k} &= \argmin_{\thetaocc} \left \| \VecA(\bm{\theta}_{\rm o})\widehat{\Vecf}^{k}  - \Vecy\right\|_2^2.
\end{align}
Moreover, by replacing $\Vecf$ in \eqref{eq:tik_reg} with its closed-form estimate, we obtain the minimization problem:
\begin{align}
    \widehat{\bm{\theta}}_{\rm o} {=} \argmin_{\thetaocc}\! \left \|\! \left[\VecA(\bm{\theta}_{\rm o})\!\left(\VecA{\!\transpose}\!(\thetaocc)\VecA\!(\thetaocc) {+} \lambda \VecI \right)\!^{-1}\VecA{\!\transpose}\!(\thetaocc) {-} \VecI\right]\!\Vecy) \right\|_2^2.
    \label{eq:occluder_minimization}
\end{align}
In principle, reconstructing $\thetaocc$ from penumbra measurement $\Vecy$ is feasible without explicitly forming estimates of $\Vecf$ as an intermediate step. We use this observation in a physics-inspired neural network inversion model outlined in \Cref{ssec:pinn}.

\subsection{Inversion via Gradient-based Optimization}

Using the uniform voxels representation (described in \Cref{ssec:voxel_uniform}), we wish to estimate $\thetaocc \defeq \{ \alpha_k\}_{k=1}^K$ and $\Vecf$ from $\Vecy$ such that \eqref{eq:discrete_forward_model} holds.
To that end, we allow the background to be non-negligible and solve the equivalent Tikhonov-regularized minimization problem:
\begin{equation}
    \argmin_{(\Vecf, \thetaocc, \Vecb, \lambda)} \left \| \VecA(\bm{\theta}_{\rm o})\Vecf + \Vecb  - \Vecy\right\|_2^2 + \lambda\|\Vecf\|_2^2.%
    \label{eq:tik_reg_bg}
\end{equation}
To impose convexity, the binary constraint is relaxed
by using $\sigma(\Vecz_k)$, where $\sigma(\cdot)$ is the sigmoid function, as a proxy for the binary-valued $\alpha_k$. Hence, we instead optimize $\Vecz_k$ using gradient descent. This approach demonstrates robust performance on simulated datasets.
In real experiments, we found that incorporating a positivity constraint when background and noise contributions are not negligible is more effective.

\Cref{alg:ASO} summarizes the steps of our voxelized optimization-based approach. 
Empirical results (see supplementary material) show superior convergence properties when directly optimizing for $\Vecf, \Vecz, \Vecb$ and $\lambda$ in an alternating fashion using \Cref{alg:ASO}, compared to assuming a negligible background.

\begin{algorithm}[h]
\caption{Alternating Minimization Method}
\label{alg:ASO}
\begin{algorithmic}[1]
\Require $ \VecA, [\VecV_1, \VecV_2, \VecV_3, ..., \VecV_K], \Vecy, \texttt{numIter} $
\State Initialize $ \Vecz_0 $ randomly, $ \Vecb_0 = \bm{0}$, and  $ \lambda_0 = 1$
\State Initialize step sizes $\eta_b$, $\eta_z$, and $\eta_\lambda$
\For{ $ i = 1 $ to $ \texttt{numIter} $}
    \State $ \VecA_v = A \odot (1 - \frac{1}{K}\sum_{k=1}^K\VecV_k \cdot \mathbb{\sigma}(\Vecz_{i_k})) $ {\hspace*{\fill}{\scriptsize Impose binary constraint }}
    \State $ \Vecf^* = (\VecA_v{\transpose}\VecA_v + \lambda_{i-1}I)^{-1} \VecA_v{\transpose}\Vecy $ \hspace*{\fill}{\scriptsize Estimate 2D non-occluding scene}
    \State  $ \Vecf_i = \max(\Vecf^* + \Vecb_{i-1}, 0) $ 
    \State $ \mathcal{L}(\Vecy, \VecA_v\Vecf_i) = \frac{1}{M} \sum_{j=1}^{M} \| \Vecy_j - (\VecA_v\Vecf_{i})_j \|_2^2 $
    \State $ \Vecb_i \leftarrow \Vecb_{i-1} - \eta_b \frac{\partial \mathcal{L}(\Vecy, \VecA_v \Vecf_i)}{\partial \Vecb_{i-1}} $ 
    \hspace*{\fill}{\scriptsize Update background contribution estimate}
    \State $ \Vecz_i \leftarrow \Vecz_{i-1} - \eta_z \frac{\partial \mathcal{L}(\Vecy, \VecA_v \Vecf_i)}{\partial \Vecz_{i-1}} $ 
    \hspace*{\fill}{\scriptsize Update occluder estimate}
    \State $ \lambda_i \leftarrow \lambda_{i-1} - \eta_\lambda \frac{\partial \mathcal{L}(\Vecy, \VecA_v \Vecf_i)}{\partial \lambda_{i-1}} $ 
    \hspace*{\fill}{\scriptsize Update estimation of the regularization constant}
\EndFor
   \State \textbf{Return} \( \widehat{\Vecf} = \Vecf_i , \widehat{\bm{\alpha}} = \sigma(\Vecz_i ) > 0.5 \)
\end{algorithmic}
\end{algorithm}
\vspace{-4.5mm}
\subsubsection{Implementation details:} The alternating minimization algorithm (\Cref{alg:ASO}) is implemented in PyTorch. The occluder region is discretized into $N_x \times N_y \times N_z$ uniform voxels.
A high discretization is required for occluders.
The required computation and memory are high, even for coarse discretizations like $10\times5\times10$.

To alleviate the high complexity and memory needed to precompute the matrix $\VecV_k$ for all possible pinspeck contributions, we employ a sparse matrix representation for $\VecV_k$.
This approach stores only the specific pixel locations to which each pinspeck contributes. 
The required gradient updates (for all pinspeck proxies $\{\Vecz_k\}_k$) are computed in this new representation. (See supplementary material for more details and simulated examples.)

\subsection{Inversion via Physics-inspired Neural Network}
\label{ssec:pinn}
Here, we propose a reconstruction approach that unifies our physics-based separable model with SSD as a learned shape prior. Specifically, because SSD generates plausible point clouds of 3D shapes alone and not their locations in the NLOS scene, we augment our estimation pipeline with an adaptation of the location estimation algorithm in \cite{saunders2019computational}.
Our entire reconstruction pipeline, which includes estimation of $\Vecf$, proceeds as follows:

\begin{itemize}
    \item {\textbf{Stage 1: Conditional generation of 3D shapes.} } Sample the 3D shape from a conditional diffusion model given the soft shadow photograph $\Vecy$.
\begin{equation}
\{\widehat{\Vecu}_k^0\}_{k=1}^K =  G_w(\Vecy),
\end{equation} 

\item {\textbf{Stage 2: Localization of generated 3D shapes.}}
Estimate the occluding object's spatial location by finding the 3D translation vector, $\widehat{\bm{\delta}_{\Vecu}}$, with:

$\widehat{\bm{\delta}_{\Vecu}} = \argmax_{\bm{\delta}_{\Vecu}} \left\| \VecH(\{\widehat{\Vecu}_k^0 + \bm{\delta}_{\Vecu}\}_{k=1}^K)  \Vecy \right\|_2^2$,
where $\VecH(\cdot) = \VecA(\cdot) \left( \VecA(\cdot)^T \VecA(\cdot) \right)^{-1} \VecA(\cdot)^T$.

\item {\textbf{Stage 3: Reconstruction of the non-occluding component. }} We convert the properly translated occluder point cloud $\{\widehat{\Vecu}_k^0 + \widehat{\bm{\delta}_{\Vecu}}\}_{k=1}^K$, and compute its mesh representation to obtain $\widehat{\bm{\theta}}_{\rm o}$. We then formulate and solve the following TV regularized problem: $\widehat{\Vecf} = \underset{\Vecf}{\mathrm{arg\,min}}||\Vecy - \VecA(\widehat{\bm{\theta}}_{\rm o})\Vecf||_2^2 + \lambda||\Vecf||_{\rm TV}^2$.

\end{itemize}
\vspace{-8mm}
\subsubsection{Implementation details:} We implement our physics-inspired neural network model, which includes the SSD network, in PyTorch.
\vspace{-3mm}
\paragraph{\textbf{Datasets.}}
We train our SSD model on thousands of 3D models. We rendered all 3D models in the ShapeNet \cite{chang2015shapenet} dataset from 3 random camera angles as
RGB images using Blender \cite{blender}. These images depict the non-occluding objects confined on the 2D plane, while the 3D models are the light-occluding structures. Our Blender script normalizes each 3D model to a bounding unit cube, configures a standard lighting setup, and exports RGB images using Blender’s built-in real-time rendering engine.
We then utilized the rendering equation in \eqref{eq:discrete_forward_model} on all possible pairs of rendered images and 3D models; and randomly selected pairs to simulate a diverse set of training examples.
To produce point clouds of a 3D model, we first constructed a dense point cloud for each object by sampling 50,000 points along the 3D object's surface. The dense point cloud of unevenly spaced points is then resampled to produce a point cloud of 2,048 evenly spaced points. Our dataset is roughly 260,000 simulated instances of various object categories.
\vspace{-3mm}
\paragraph{\textbf{Soft Shadow Image Encoder.}}
To encode the soft shadow photographs into latent representation, $\hat{\Vecy}$, we employed the CrossFormer architecture~\cite{crossformer} and trained it jointly with the diffusion process. CrossFormer introduces a cross-scale attention mechanism crucial for processing visual inputs. In addition, unlike traditional vision transformers, CrossFormer effectively captures the correlation between features of different scales, which is vital for understanding complex visual inputs like our penumbra photographs.
\vspace{-3mm}
\paragraph{\textbf{Model Implementation.}}
Deviating from previous work on point cloud reconstructions that often employed specialized 3D architectures~\cite{zhou20213d} or transformer-based~\cite{nichol2022pointe} models, we take advantage of a simpler yet effective UNet-based architecture \cite{Ronneberger2015UNetCN} with 1D convolutions. Our model allows fast sampling of more points than the model was trained on, and is designed to predict the mean $\mu$ of the noise in \eqref{eq:loss_function}, from the inputs
$\hat{\Vecy}$, $t$, and $\{\Vecu^t_k\}_{k=1}^K$, i.e., the encoded representation of a soft shadow 2D photograph, current time step $t$ during the diffusion process, and the intermediate noisy point cloud, respectively. The entire model architecture is illustrated in \Cref{fig:ssd_model} and additional details are provided in the supplementary document.
\vspace{-3mm}
\paragraph{\textbf{3D mesh representation.}} 
Reconstructing the non-occluding portion of the hidden scene $\Vecf$ requires accurate rendering of soft shadow images of the estimated occluding objects. Sparse point clouds may lead to inaccuracies that yield poor approximations of the soft shadows.
Hence, we convert the generated point clouds from \textit{soft shadow diffusion} into meshes. Using the entire ShapeNet \cite{chang2015shapenet} dataset, we train a signed distance field (SDF) neural network model to predict the SDF of the reconstructed point cloud. The SDF model follows the autoencoder architecture~\cite{chou2022diffusionsdf}. We then apply the marching cubes algorithm~\cite{LorensenC87} to the computed SDF to generate a mesh representation of the light-occluding object.
The mesh representations facilitate higher fidelity renders of the soft shadows.

\section{Experiments}
\label{sec:Results}
Experimental results for real and synthetic data are presented to demonstrate the efficacy of our proposed methods.
Because prior methods focus on different acquisition scenarios (such as access to video frames~\cite{Yedidia_2019_CVPR,yedidia2019using}, pre-calibrated occluder visibility functions~\cite{Baradad2018inferring}, known occluder shapes~\cite{saunders2019computational}, or even fully known occluders~\cite{Geng2022passive,Wang2021accurate}) direct and fair comparisons are not possible. These prior techniques focus on reconstructing the light-emitting portion of the hidden scene alone, whereas our approach is the first to seek, in addition, a 3D reconstruction of hidden occluders. We are considering a much harder problem and as such the 2D reconstructions achievable will, in expectation, be no better than existing methods.

\begin{figure*}[ht!]
\vspace{-8mm}
\centering
\begin{subfigure}[b]{\textwidth}
    \includegraphics[width=\textwidth]{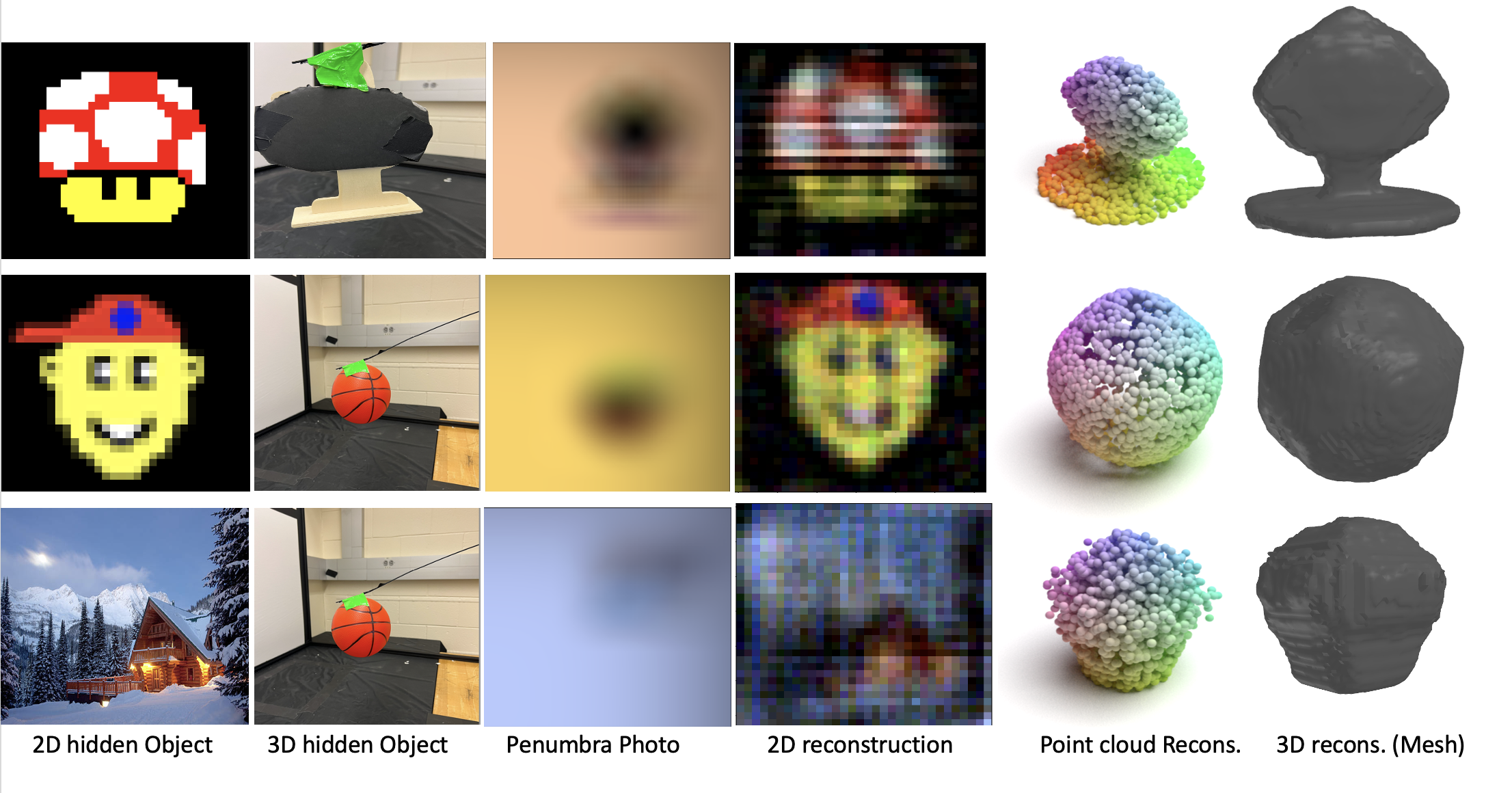}
    \caption{SSD-based reconstruction on Simple 3D Objects.}
    \label{fig:A}
\end{subfigure} \\
\begin{subfigure}[b]{\textwidth}
    \includegraphics[width=\textwidth]{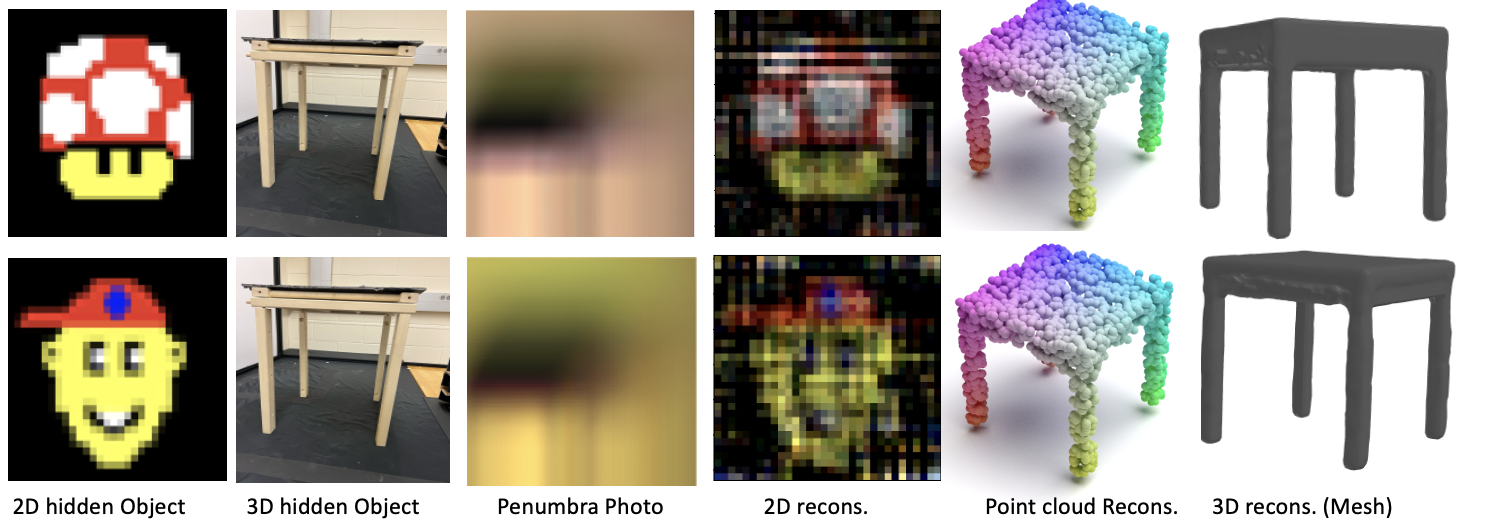}
    \caption{SSD-based reconstruction on Large 3D Objects.}
    \label{fig:B}
\end{subfigure}
\caption{\textbf{Real Experimental Reconstruction Results from Soft Shadow Diffusion Model.}
The model takes the soft shadow photographs (column 3) as input to condition the diffusion process and generates a 3D mesh representation (column 6) of the 3D light-occluding structure (column 2) in the hidden area. The mesh representation is then translated for the reconstruction of the light-emitting objects (column 4) as a 2D RGB image.}
\label{fig:real_ssd_results}
\end{figure*}
\begin{figure}[h]
\centering
\begin{subfigure}[b]{0.4\textwidth}
    \includegraphics[width=\textwidth]{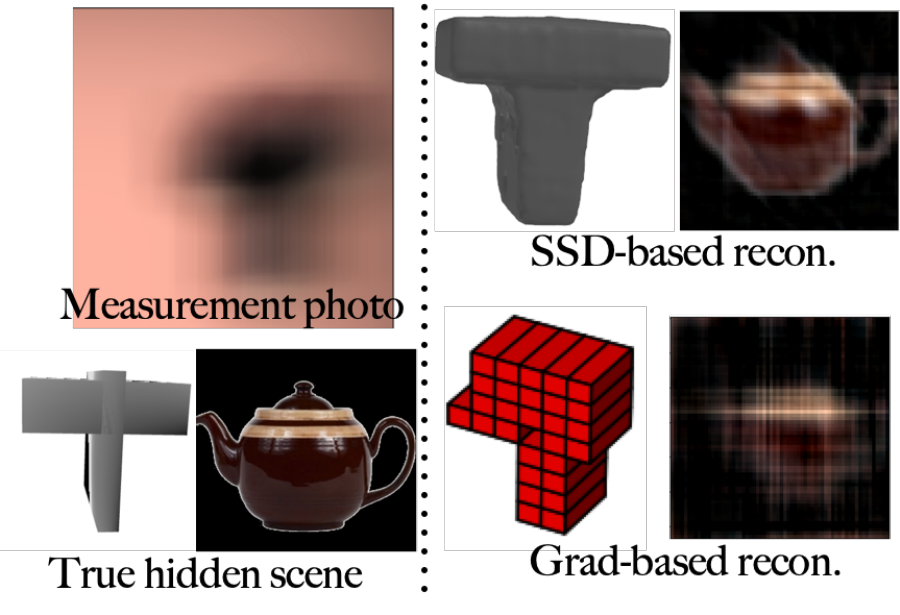}
    \caption{Simulated data.}
    \label{fig:sim_recon_comp}
\end{subfigure}
\hfill
\begin{subfigure}[b]{0.5\textwidth}
    \includegraphics[width=\textwidth]{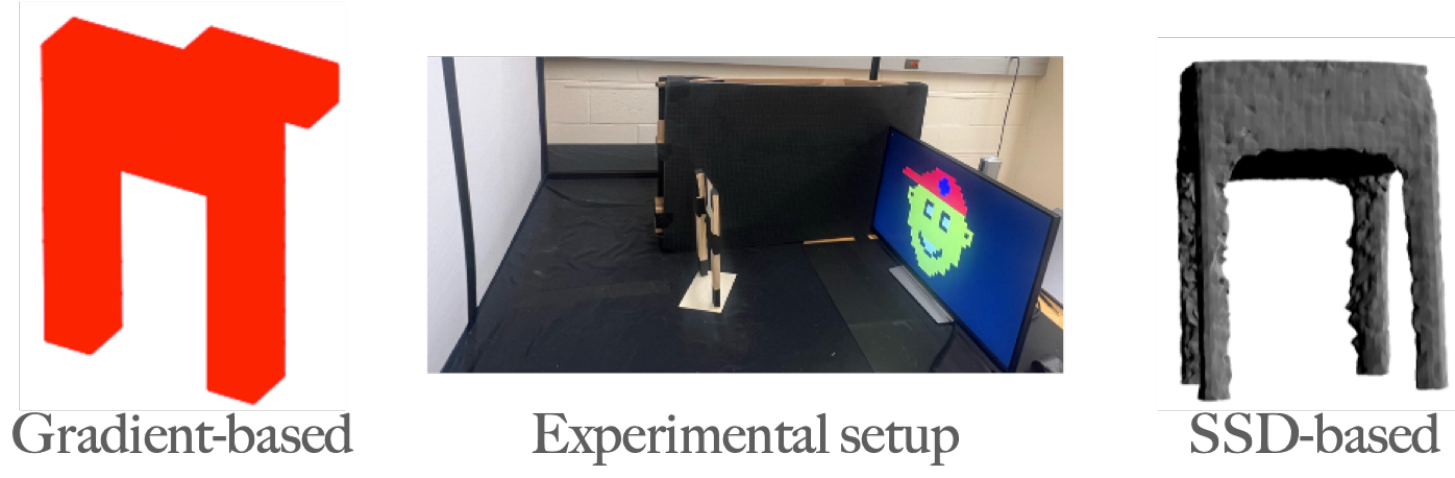}
    \caption{Real data.}
    \label{fig:real_recon_comp}
    \vspace{9mm}
\end{subfigure}
\vspace{-4mm}
\caption{{Comparing reconstructions of SSD and gradient-based optimization.} SSD is trained on 3D objects and outputs a (3D) four-legged table for the planner occluder. 
The gradient-based reconstructions use $10\times5\times10$ discretization of the occluding region.}
\label{fig:real_physics_results}
\vspace{-5mm}
\end{figure}
\vspace{-8mm}
\subsection{Qualitative Results}
Real experimental examples of the \textit{soft shadow diffusion} are presented in \Cref{fig:nlos_config,fig:real_ssd_results}, and the real experimental examples of the \textit{gradient-based} method are shown in \Cref{fig:real_physics_results}. \Cref{fig:real_ssd_results,fig:real_physics_results} depicts the non-occluding portion of the hidden scene on a monitor at a distance of 1.08 meters to the visible wall, and 3D occluders placed between them. Measurements were obtained by photographing the visible wall with a 4-megapixel camera. The measurements were inputs to our physics-inspired neural network inversion method or Algorithm \ref{alg:ASO} for the gradient-based inversion method. In the final stage of SSD, we reconstruct each channel of the 2D non-occluding portion using TV regularization.
The reconstructions show good qualitative agreement with the ground truth hidden scenes. However, the gradient-based method does not scale well for more complex occluder shapes requiring finer discretizations (as shown in the supplementary document).

Even when the non-occluding hidden scene portion reflects light (rather than being self-luminous) and spans a large depth (instead of being confined to a plane), the reconstructed 3D image of the occluder and 2D image of the non-occluders are remarkably accurate as shown in \Cref{fig:nlos_config}. This experiment further highlights the surprising generalizability of the proposed SSD-based method, which was trained using only simulated data.

Additional qualitative results with real experiments and simulations are given in the supplement.

\begin{figure}
    \centering
    \begin{subfigure}{.95\textwidth}
        \centering
        \includegraphics[width=0.75\linewidth]{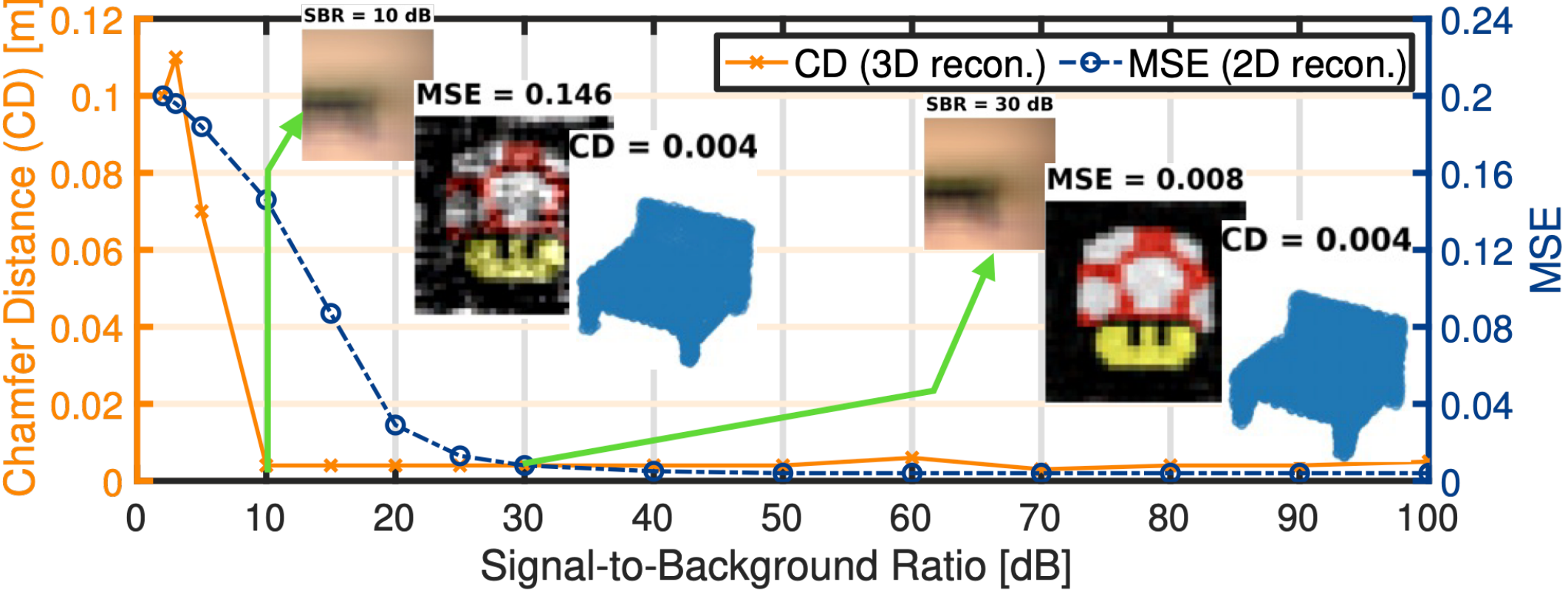}
        \caption{Reconstructions from simulated data.}
    \label{fig:sbr_plot_ssd_s}
    \end{subfigure}%
    \\
    \begin{subfigure}{.95\textwidth}
        \centering
        \includegraphics[width=1\linewidth]{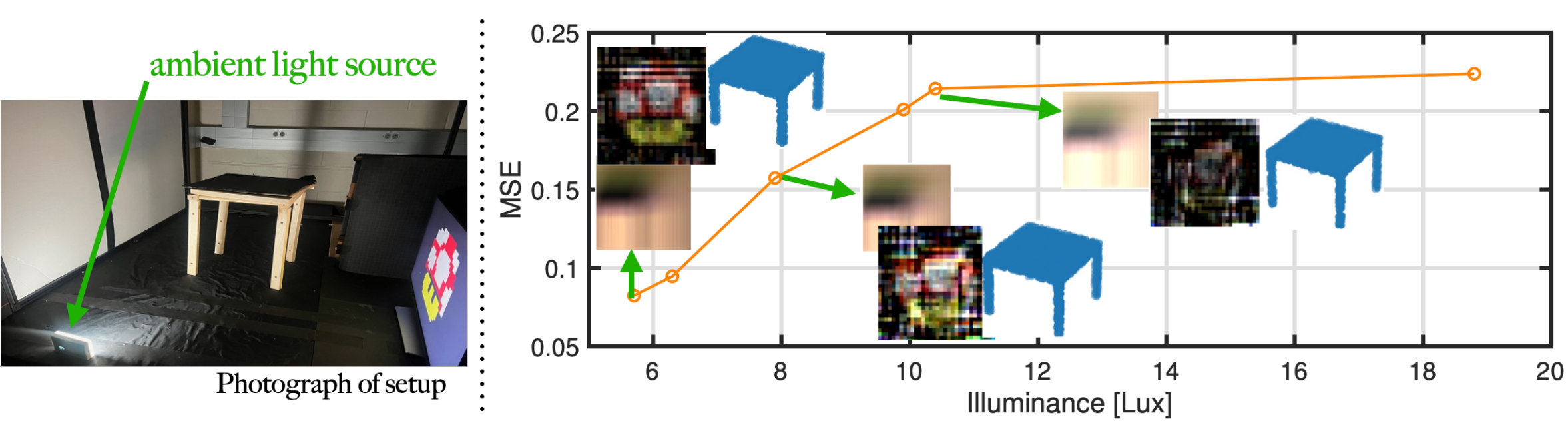}
        \caption{Reconstructions from real experimental data.}
    \label{fig:sbr_plot_ssd_s2}
    \end{subfigure}%
    \caption{\textbf{Robustness of SSD-based approach to background illumination.} The insights show the measured photograph and reconstructions at each background level. With the monitor brightness fixed, a larger illuminance value indicates increased ambient light source contributions in (b).}
    \label{fig:quantitative}
\vspace{-5mm}
\end{figure}
\subsection{Quantitative Results}
Here, we investigate the robustness of our approach to increasing levels of background illumination (\Cref{fig:quantitative}), while reconstructions obtained under varying levels of measurement noise are presented in the supplementary document (Supplementary Section 4.1). In the real experimental result, we introduce an additional external light source that increases the overall brightness of the captured photograph but does not cast penumbra.
We report the mean squared error (MSE) for the 2D reconstructions, and Chamfer Distance (CD) for the 3D occluding object reconstructions. Our approach degrades slowly for reasonable SBRs in simulated experiments, with the 3D reconstructions remaining accurate even in SBRs of around 10\,dB, 2D reconstructions degrade gradually for SBRs below 20\,dB\@. Similar trends are observed for real data reconstructions. The surprising stability of the 3D reconstructions even at very low SBRs demonstrates the utility of the strong shape prior imposed by the proposed SSD network.

SSD demonstrates similar robustness to measurement noise, with 3D reconstruction quality relatively unaffected even at signal-to-noise ratios as low as 10\,dB\@. Additional experiments, shown in Supplementary Section 4.2, suggest that SSD also generalizes well to reconstructing unseen 3D shape classes.

\section{Discussion \& Conclusion}
We introduced a formulation of computational periscopy that enables 3D reconstructions. Our formulation inspired a physics-based algorithm, and a physics-inspired neural network for computing 3D reconstructions of light-occluders, in addition to 2D reconstructions of the light-reflecting structures in the hidden scene.
This demonstrates an advance over prior techniques in occluder-aided passive NLOS imaging that relied on \textit{a priori} known information about the light-occluders in the scene.
Central to our physics-inspired neural network model is a novel soft shadow diffusion (SSD) network, which reconstructs point clouds from soft shadows and extends denoising diffusion models to interpret soft shadows. Although SSD was trained in simulations, both approaches were successful in real experiments. SSD was further shown to degrade gracefully with noise and additional background.

Notwithstanding, some aspects of our work warrant further investigation: First, our gradient-based method is limited to reconstructing simple shapes due to the computational demand of higher discretization. Although a sparse matrix approach mitigated this limitation in simulations, it was less effective in real-world scenarios (as shown in the supplementary material). Second, our SSD model predicts a view-consistent 3D point cloud without information about the true location of the points in the NLOS configuration. To facilitate recovery of the non-occluding object, we partition the occluder space into multiple 3D bounding boxes of the occluder size. The subsequent step involves a grid search over these bounding boxes to position the occluder accurately within the hidden area. However, the size of the occluding object is typically unknown in real-world uncontrolled scenarios. These shortcomings are avenues for future research.
\bibliographystyle{splncs04}
\bibliography{main}
\newpage
\appendix
\onecolumn
\def\Veca{\mathbf{a}}
\def\Vecb{\mathbf{b}}
\def\Vecc{\mathbf{c}}
\def\Vecd{\mathbf{d}}
\def\Vecf{\mathbf{f}}
\def\Vech{\mathbf{h}}
\def\Veck{\mathbf{k}}
\def\Vecn{\mathbf{n}}
\def\Vecp{\mathbf{p}}
\def\Vecq{\mathbf{q}}
\def\Vecr{\mathbf{r}}
\def\Vecs{\mathbf{s}}
\def\Vecu{\mathbf{u}}
\def\Vecv{\mathbf{v}}
\def\Vecw{\mathbf{w}}
\def\Vecx{\mathbf{x}}
\def\Vecy{\mathbf{y}}
\def\Vecz{\mathbf{z}}

\def\Veckappa{\bm{\kappa}}
\def\Vecphi{\bm{\phi}}

\def\VecA{\mathbf{A}}
\def\VecB{\mathbf{B}}
\def\VecC{\mathbf{C}}
\def\VecD{\mathbf{D}}
\def\VecE{\mathbf{E}}
\def\VecF{\mathbf{F}}
\def\VecG{\mathbf{G}}
\def\VecI{\mathbf{I}}
\def\VecJ{\mathbf{J}}
\def\VecK{\mathbf{K}}
\def\VecH{\mathbf{H}}
\def\VecM{\mathbf{M}}
\def\VecP{\mathbf{P}}
\def\VecR{\mathbf{R}}
\def\VecT{\mathbf{T}}
\def\VecU{\mathbf{U}}
\def\VecV{\mathbf{V}}
\def\VecW{\mathbf{W}}
\def\VecX{\mathbf{X}}
\def\VecY{\mathbf{Y}}
\def\VecZ{\mathbf{Z}}
\def\thetaocc{\bm{\theta}_{\rm o}}


\def\transpose{^{\!\mathsf{T}}}
\def\hermitian{^{\!\mathsf{H}}}
\def\gfunc{\textit{g}}
\def\ffunc{\textit{f}}
\def\vfunc{\textit{v}}





\title{Supplementary Material for \\
``Soft Shadow Diffusion (SSD): Physics-inspired Learning for
3D Computational Periscopy''}

\titlerunning{Supplement-Soft Shadow Diffusion (SSD)}

\author{Fadlullah Raji\orcidlink{0009-0003-0619-0562} \and
John Murray Bruce\orcidlink{0000-0002-1416-4175}}

\authorrunning{Raji and Murray-Bruce}

\institute{ University of South Florida.\\ 4202 E. Fowler Avenue, Tampa, FL, USA 33620.  \\
{\tt\small fraji@usf.edu }
{\tt\small murraybruce@usf.edu} \\}

\maketitle

This supplemental document to our manuscript provides additional information on the complexity of the purely physics-based approach (\Cref{sup_sec:physics_complexity}) and its robustness to unknown ambient illumination (\Cref{sup_sec:physics_bg}).
Additional details on implementing and training our physics-inspired soft shadow diffusion model are provided in \Cref{sup_sec:ssd}. Finally, \Cref{sup_sec:expt} contains various additional real and synthetic experimental results. It explores the stability of the SSD model to noise and background illumination and also shows additional 3D reconstructions.

\section{Physics-based 3D Computational Pericopy: Towards higher resolution voxelizations}
\label{sup_sec:physics_complexity}
Figure 5(b) in the main paper shows a reconstruction from real experimental data using a $10 \times 5 \times 10$ voxels discretization of the \textit{occluding region}.
(Note that the portion of the hidden scene which we refer to as the \textit{occluding region} is illustrated in Figure 2 of the main manuscript.)

\subsection{Complexity Analysis of Physics-based Reconstruction}
The $10 \times 5 \times 10$~pinspecks voxelization is relatively coarse to be able to describe a large variety of 3D objects. Increasing the number of pinspecks comes with increased computational complexity and space requirements. Figure 5(a) in the main manuscript is an example of increasing the grid size which we reconstructed using $16 \times 5 \times 16$ discretization of the occluding region. Given a camera photograph of $128 \times 128$, and the non-occluding scene to be estimated at $32 \times 32$ resolution, the memory requirement will be $16384 \times 1024 \times 1280$ = 80GB when using Float32 precision. Since our algorithm requires the computation of gradient and updates of the variables to be estimated, this memory requirement doubles. Hence, to achieve the simulated reconstruction in  Figure 5(a) of the main manuscript, we used 3 A100 GPU with 80GB each, and distribute the  forward model across these GPUs for computation. A sparse representation of this forward model is presented next.

\subsection{Sparse Matrix Representation}
\begin{figure}[tbh!]
    \centering
    \begin{subfigure}{.3\linewidth}
        \centering
        \includegraphics[width=0.8\linewidth]{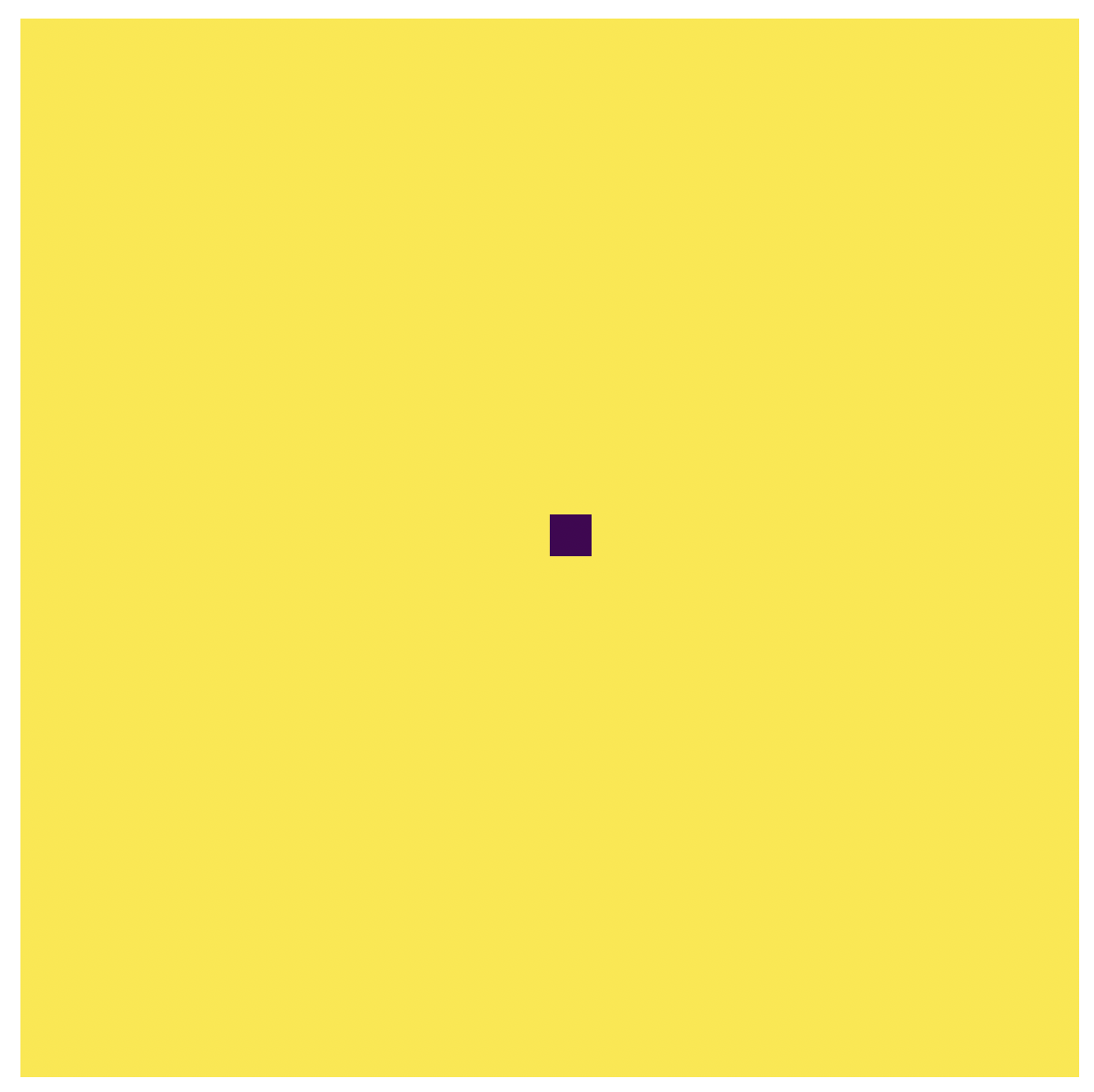}
        \label{fig:pinspeck-contribution1}
    \end{subfigure}
    \begin{subfigure}{.3\linewidth}
        \centering
        \includegraphics[width=0.8\linewidth]{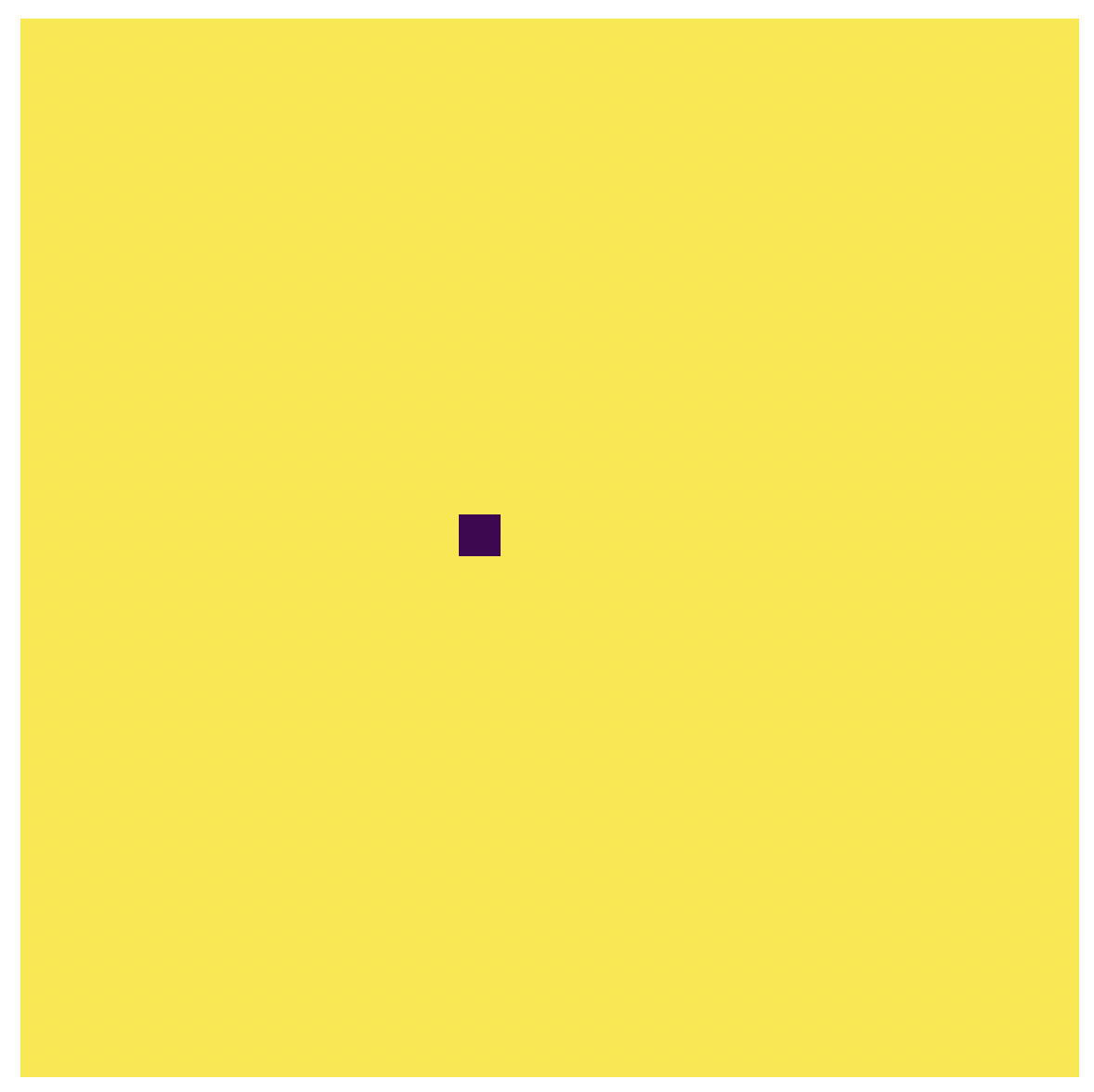} 
        \label{fig:pinspeck-contribution2}
    \end{subfigure}
    \begin{subfigure}{.3\linewidth}
        \centering
        \includegraphics[width=0.8\linewidth]{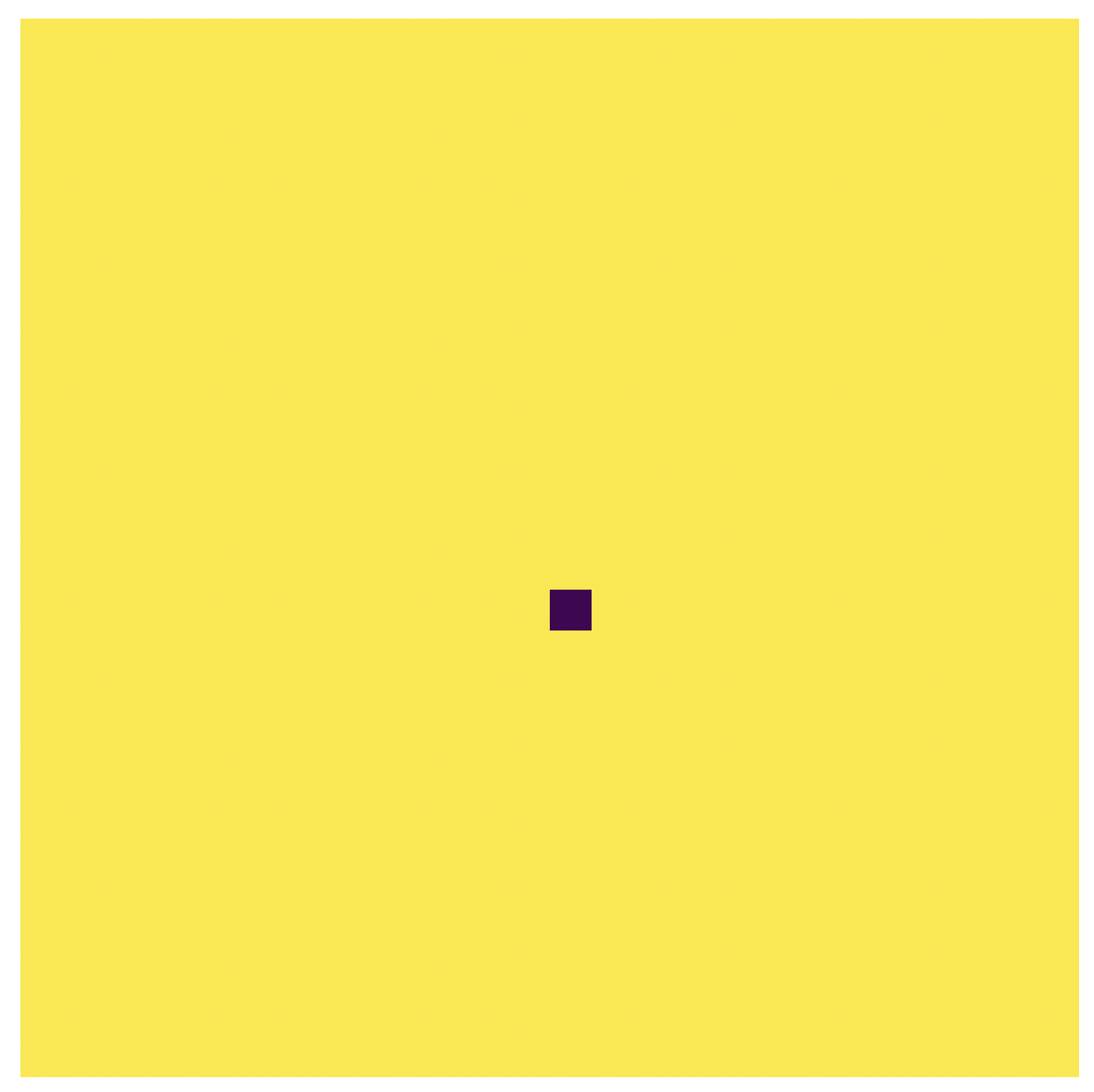} 
        \label{fig:pinspeck-contribution3}
    \end{subfigure}
     \caption{Each panel shows the variation of the visibility function over the visible wall/camera FOV for some patch in the hidden-scene light-emitting 2D plane and a hidden-scene pinspeck voxel. The visibility function is binary-valued, with one (yellow) indicating that the patch in the hidden-scene light-emitting 2D plane is unoccluded from those patches on the visible wall, and zero (dark blue) indicating occlusion by the pinspeck voxel.
     }
     \label{fig:pinspeck-contribution}
\end{figure}

The contribution (shadow) of a hidden scene pinspeck due to a point light source in the hidden scene may be interpreted using a sparse structure for $\VecV_k$. Here, the contribution is zero at a few locations and one everywhere else, as shown in \Cref{fig:pinspeck-contribution}. More precisely, the one's complement of the contribution is sparse.
Thus, to achieve higher resolution, the discrete forward model in main manuscript equation (10) can be evaluated without saving the large matrix $\VecV_k$  for each pinspeck location indexed by $k$.
Instead, a sparse data structure may be utilized to improve efficiency.

Results of adopting a sparse representation in evaluating the pinspeck forward model at higher reconstruction resolutions are shown in \Cref{fig:pinspeck-contribution}.
Although we are able to reconstruct correctly in simulation, accurate real-world reconstruction is still challenging.
This is, in part, due to the pinspeck approximation. Notwithstanding this shortcoming, this physics-based formulation provides a flexible framework that enables our physics-inspired neural network approach. Our SSD model can handle higher discretizations even for real experimental datasets, as shown in results presented in the main manuscript (Figures 5) and in Supplementary \Cref{fig:real_ssd} (shown here).

\begin{figure}[htb!]
    \centering
    \includegraphics[width=0.9\linewidth]{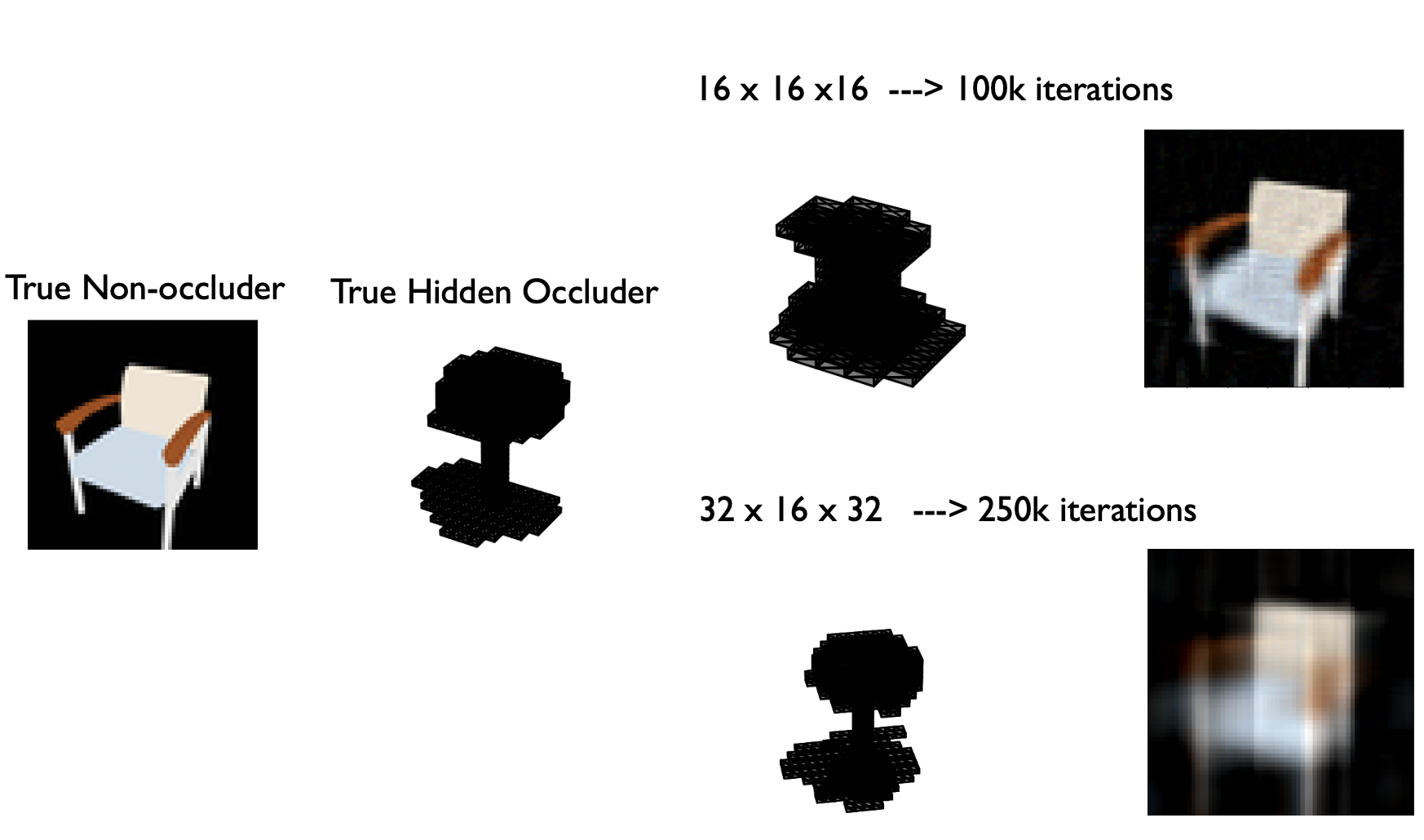}
    \caption{Increased resolution reconstructions in simulations using sparse data structures.
    }
    \label{fig:physics-high-resolution}
\end{figure}

\clearpage
\section{Gradient-based inversion: Neglecting background vs Modeling background contributions}
\label{sup_sec:physics_bg}
\begin{figure}[ht!]
    \centering
    \begin{subfigure}[t]{0.5\textwidth}
        \centering
        \includegraphics[width=0.9\linewidth,trim={0cm 0 12cm 0},clip]{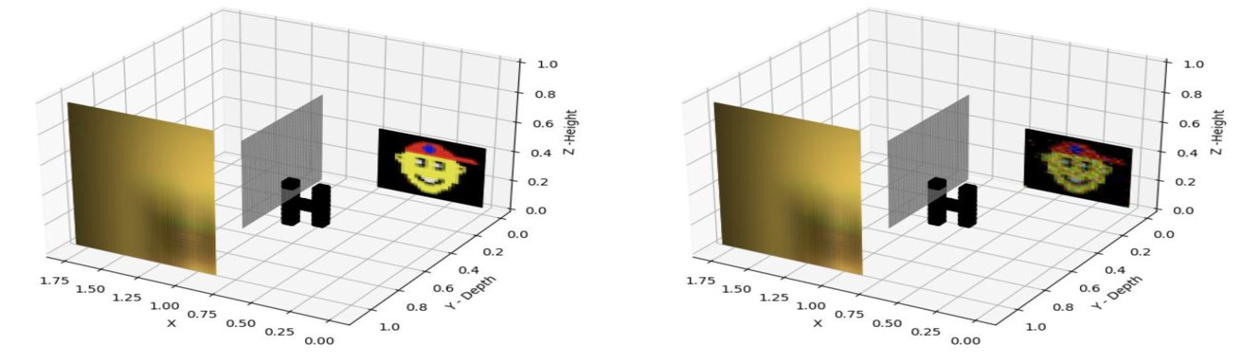}
        \caption{Depiction of ground truth scene.}
    \end{subfigure}
    \\
    \begin{subfigure}[t]{0.5\textwidth}
        \centering
        \includegraphics[width=0.9\linewidth,trim={12cm 0 0cm 0},clip]{suppFigures/gradient-based.png}
        \caption{Jointly estimating background contributions.}
    \end{subfigure}%
    ~ 
    \begin{subfigure}[t]{0.5\textwidth}
        \centering
        \includegraphics[width=0.9\linewidth,trim={9cm 0 0cm 0},clip]{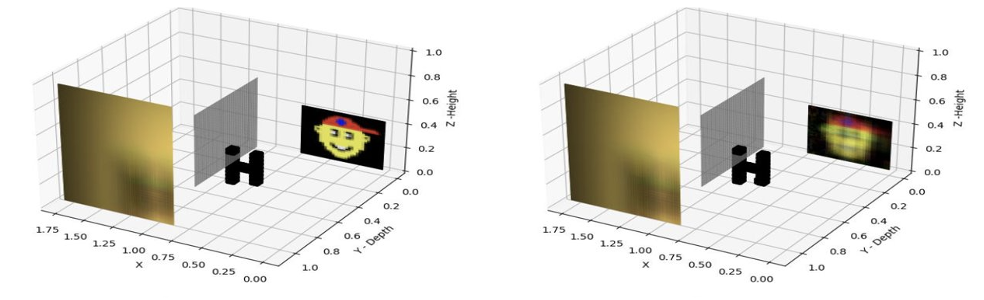}
        \caption{Without estimating background.}
    \end{subfigure}
    \caption{\textbf{Reconstructions from gradient-based inversion via alternating minimization.}
    Panels (b) and (c) show the reconstructions of the 3D occluding structures and of the 2D light-emitting scene, with (b) and without (c) incorporating background modeling and estimation.
    }
    \label{fig:Gradient-Based_nob_vs_b_solution}
\end{figure}

\begin{figure}[htb!]
\centering
    \includegraphics[width=1\linewidth]{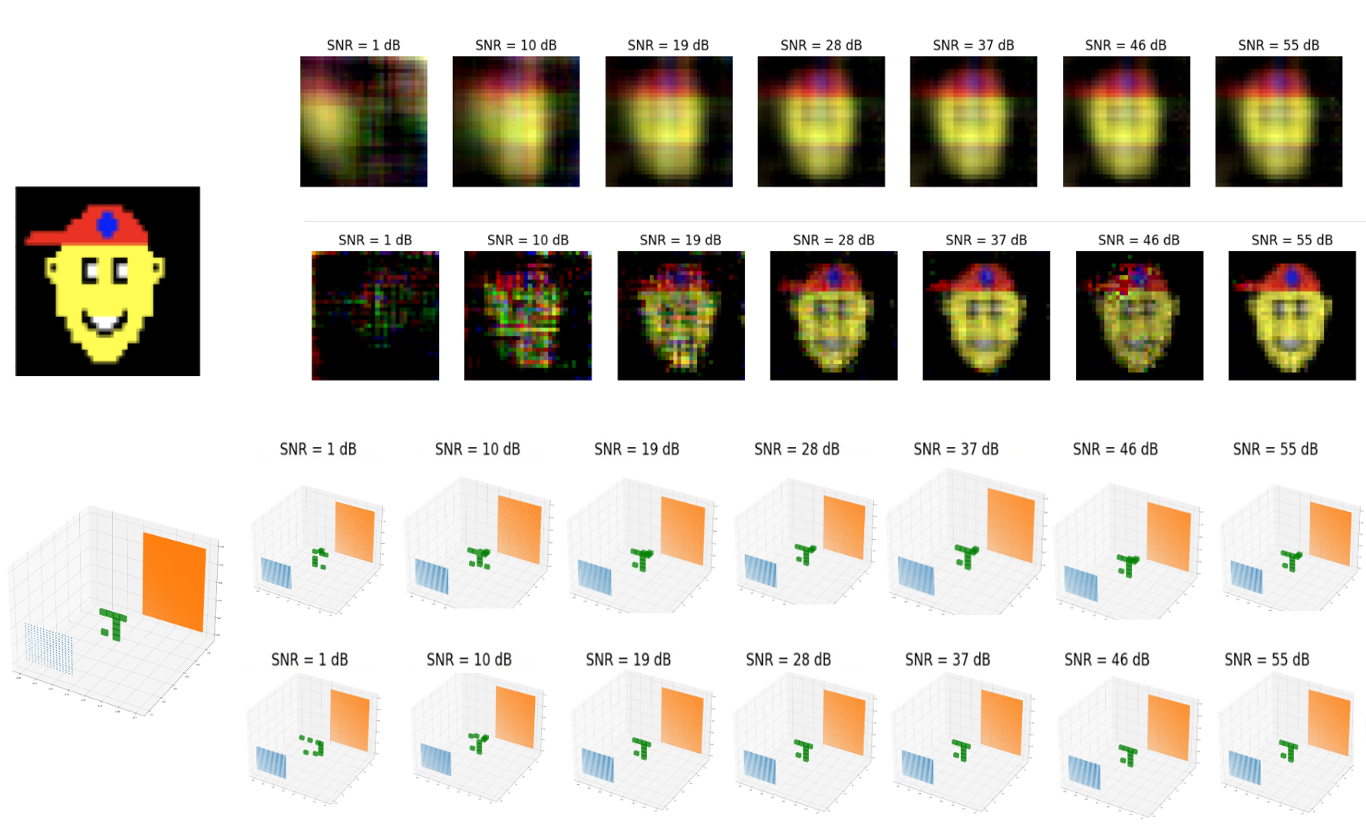}
    \caption{\textbf{Comparing the two variants of the gradient-based hidden scene reconstruction algorithm.}
    Row 1: 2D reconstruction of the light-emitting hidden scene component when background contribution $\Vecb$ is neglected.
    Row 2: 2D reconstruction of the light-emitting hidden scene component when background contribution $\Vecb$ is modeled and estimated.
    Row 3: 3D reconstruction of the light-occluding hidden scene component when background contribution $\Vecb$ is neglected.
    Row 4: 3D reconstruction of the light-occluding hidden scene component when background contribution $\Vecb$ is modeled and estimated.}
    \label{fig:GradientBenchmark}
\end{figure}

We compare the gradient-based inversion via alternating minimization, and a variant that incorporates optimization of the background contribution in \Cref{fig:Gradient-Based_nob_vs_b_solution,fig:GradientBenchmark}. It shows that the solution without learning the noise, and background contribution is blurry. Optimizing the background noise is more efficient, and produces sharp image of the non-occluding structure.

\begin{algorithm}[h]
\caption{Alternating Minimization Method with Background Neglected}
\label{alg:ASO_variant}
\begin{algorithmic}[1]
\Require $ \VecA, [\VecV_1, \VecV_2, \VecV_3, ..., \VecV_K], \Vecy, \texttt{numIter} $
\State Initialize $ \Vecz_0 $ randomly, 
and  $ \lambda_0 = 1$
\State Initialize step sizes 
$\eta_z$, and $\eta_\lambda$
\For{ $ i = 1 $ to $ \texttt{numIter} $}
    \State $ \VecA_v = 1 - \frac{1}{K}\sum_{k=1}^K\VecV_k \cdot \mathbb{\sigma}(\Vecz_{i_k}) $
    \State $ \Vecf^* = (\VecA_v{\transpose}\VecA_v + \lambda_{i-1}I)^{-1} \VecA_v{\transpose}\Vecy $    
    \State $\Vecf_i = \max(\Vecf^*, 0)$
    \State $ \mathcal{L}(\Vecy, \VecA_v\Vecf_i) = \frac{1}{M} \sum_{j=1}^{M} \| \Vecy_j - (\VecA_v\Vecf_{i})_j \|_2^2 $ 
    \State $ \Vecz_i \leftarrow \Vecz_{i-1} - \eta_z \frac{\partial \mathcal{L}(\Vecy, \VecA_v \Vecf_i)}{\partial \Vecz_{i-1}} $
    \State $ \lambda_i \leftarrow \lambda_{i-1} - \eta_\lambda \frac{\partial \mathcal{L}(\Vecy, \VecA_v \Vecf_i)}{\partial \lambda_{i-1}} $
\EndFor
   \State \textbf{Return} \( \widehat{\Vecf} = \Vecf_i , \widehat{\bm{\alpha}} = \sigma(\Vecz_i ) > 0.5 \)
\end{algorithmic}
\end{algorithm}

\clearpage
\section{Soft Shadow Diffusion Model}
\label{sup_sec:ssd}
Denote the pinspeck point cloud by $\{\Vecu_k\}_{k=1}^K = \hat{\Vecu}$. The joint distribution $p_{\theta}(\hat{\Vecu}^{(0:T)})$ is referred to as the reverse process. This process is characterized as a Markov chain with learned Gaussian transitions initiating from $p(\hat{\Vecu}^{(T)}) = \mathcal{N}(\hat{\Vecu}^{(T)}; \mathbf{0}, \mathbf{I})$:
\begin{align}
p_{\theta}(\hat{\Vecu}^{(0:T)}) := p(\hat{\Vecu}^{(T)})\prod_{t=1}^{T} p_\theta(\hat{\Vecu}^{(t-1)}|\hat{\Vecu}^{(t)}),
\\
\quad p_{\theta}(\hat{\Vecu}^{(t-1)}|\hat{\Vecu}^{(t)}) := \mathcal{N}(\hat{\Vecu}^{(t-1)}; \mu_{\theta}(\hat{\Vecu}^{(t)}, t), \sigma^2\mathbf{I}).
\end{align}

The approximate posterior $q(\hat{\Vecu}^{1:T}|\hat{\Vecu}^0)$, termed the forward process is defined as a Markov chain that incrementally incorporates Gaussian noise into the data following a predetermined variance schedule $\beta_1, \ldots, \beta_T$:
\begin{align}
q(\hat{\Vecu}^{(1:T)}|\hat{\Vecu}^0) := \prod_{t=1}^{T} q(\hat{\Vecu}^{(t)}|\hat{\Vecu}^{(t-1)}), \\
q(\hat{\Vecu}^{(t)}|\hat{\Vecu}^{(t-1)}) := \mathcal{N}(\hat{\Vecu}^{(t)}); \sqrt{1 - \beta_t} \hat{\Vecu}^{(t-1)}, \beta_t \mathbf{I})
\end{align}

Let $\hat{\Vecu} = \{\Vecu_k\}_{k=1}^K$, then the training objective is given as:
\begin{align}
\begin{split}
L(\theta, \phi) = \mathbb{E}_{q} \Bigg[ & \sum_{t=2}^{T} \sum_{k=1}^{K} \underbrace{\text{D}_{\text{KL}}\big(q(\Vecu_k^{(t-1)}|\Vecu_k^{(t)}, \Vecu_k^{(0)}) \| p_{\theta}(\Vecu_k^{(t-1)}|\Vecu_k^{(t)}, \hat{\Vecy})\big)}_{L_k^{(t-1)}} \\
& - \sum_{k=1}^{K} \underbrace{\log p_{\theta}(\Vecu_k^{(0)}|\Vecu_k^{(1)}, \hat{\Vecy})}_{L_k^{(0)}} \\
& + \underbrace{\text{D}_{\text{KL}}\big(q_{\phi}(\hat{\Vecy}|\Vecu^{(0)}) \| p(\hat{\Vecy})\big)}_{L_{\hat{\Vecy}}} \Bigg].
\end{split}
\end{align}

\subsection{Simplified Training Algorithm}
We adapt the simplified algorithm presented in \cite{ho2020denoising} to train our model. To evaluate $L_k^{(t-1)}$, we need to sample \(\Vecu_k^{(t)}\) from $q(\hat{\Vecu}^{(t)}|\hat{\Vecu}^{(0)})$. In principle, it can be done by sampling iteratively through the Markov chain. However, \cite{ho2020denoising} showed that \(q(\hat{\Vecu}^{(t)}|\hat{\Vecu}^{(0)})\) is Gaussian, thus allowing us to sample $\hat{\Vecu}^{(t)}$ efficiently without iterative sampling.
To see this, note that:
\begin{equation}
q(\hat{\Vecu}^{(t)}|\hat{\Vecu}^{(0)}) = \mathcal{N}(\hat{\Vecu}^{(t)}|\sqrt{\bar{\alpha}_t}\hat{\Vecu}^{(0)}, (1 - \bar{\alpha}_t)I).
\end{equation}
 Since both \(q(\Vecu_k^{(t-1)}|\Vecu_k^{(t)}, \Vecu_k^{(0)})\) and \(p_{\theta}(\Vecu_k^{(t-1)}|\Vecu_k^{(t)}, \hat{\Vecy})\) are Gaussians, the term \(L_k^{(t-1)}\) can be expanded as:
\begin{equation}
L_k^{(t-1)} = \mathbb{E}_{\Vecu_k^{(0)}, \Vecu_k^{(t)}, \hat{\Vecy}} \left[ \frac{1}{2\beta_t} \left\| \frac{\sqrt{\bar{\alpha}_{t-1}\beta_t}}{1 - \bar{\alpha}_t} \Vecu_k^{(0)} + \frac{\sqrt{\alpha_t (1 - \bar{\alpha}_{t-1})}}{1 - \bar{\alpha}_t} \Vecu_k^{(t)} - \mu_{\theta}(\Vecu_k^{(t)}, t, \hat{\Vecy}) \right\|^2 \right] + C.
\end{equation}

Using the Gaussian above, \(\Vecu_k^{(t)} = \sqrt{\bar{\alpha}_t}\Vecu_k^{(0)} + \sqrt{1 - \bar{\alpha}_t}\epsilon\), where \(\epsilon \sim \mathcal{N}(0, I)\):
\begin{equation}
L_k^{(t-1)} = \mathbb{E}_{\Vecu_k^{(0)}, \epsilon, \hat{\Vecy}} \left[ \frac{1}{2\beta_t} \left\| \frac{1}{\sqrt{\alpha_t}} \left( \Vecu_k^{(t)} - \beta_t \sqrt{1 - \bar{\alpha}_t} \epsilon \right) - \mu_{\theta}(\Vecu_k^{(t)}, t, \hat{\Vecy}) \right\|^2 \right] + C. \quad (14)
\end{equation}
Shown above \(\mu_{\theta}(\Vecu_k^{(t)}, t, \hat{\Vecy})\) must predict \(\frac{1}{\sqrt{\alpha_t}} \left( \Vecu_k^{(t)} - \beta_t \sqrt{1-\bar{\alpha}_t} \epsilon \right)\) given \(\Vecu_k^{(t)}\). Thus, \(\mu_{\theta}(\Vecu_k^{(t)}, t, \hat{\Vecy}))\) can be parameterized as:
\begin{equation}
\mu_{\theta}(\Vecu_k^{(t)}, t, \hat{\Vecy}) = \frac{1}{\sqrt{\alpha_t}} \left( \Vecu_k^{(t)} - \beta_t \sqrt{1 - \bar{\alpha}_t} \epsilon_{\theta}(\Vecu_k^{(t)}, t, \hat{\Vecy}) \right), \quad 
\end{equation}
where \(\epsilon_{\theta}(\Vecu_k^{(t)}, t, \hat{\Vecy})\) is the physics-inspired neural network intended to predict \(\epsilon\) from \(\Vecu_k^{(t)}\). Finally, \(L_k^{(t-1)}\) can be simplified as
\begin{equation}
L_k^{(t-1)} = \mathbb{E}_{\Vecu_k^{(0)}, \epsilon, \hat{\Vecy}} \left[ \frac{\beta_t^2}{2\beta_t\alpha_t(1 - \bar{\alpha}_t)} \left\| \epsilon - \epsilon_{\theta}(\sqrt{\bar{\alpha}_t}\Vecu_k^{(0)} + \sqrt{1 - \bar{\alpha}_t}\epsilon, t, \hat{\Vecy}) \right\|^2 \right] + C. \quad (14)
\end{equation}
To minimize \(L_i^{(t-1)}\), we can only minimize \(\mathbb{E}[\|\epsilon - \epsilon_{\theta} \|^{2}]\) because the coefficient \(\frac{\beta_t^2}{2\beta_t\alpha_t(1-\bar{\alpha}_t)}\) is constant.

The simplified algorithm \cite{ho2020denoising} proposed selecting a random term from the set $\left\{\sum_{k=1}^{K} L_k^{(t-1)}\right\}_{t=1}^{T}$ for optimization in each step of the training process.

Then the training algorithm is presented in \Cref{alg:training_ssd}.

\begin{algorithm}[h]
\caption{SSD Training algorithm}
\begin{algorithmic}[1]
\Repeat
    \State Sample the occluding components $\{\Vecu_k^{(0)}\}_{k=1}^K \sim p_{\text{data}}(\Vecu^{(0)})$
    \State Sample the non-occluding components $\Vecf \in \mathbb{R}^N \sim p_{\text{data}}(\Vecf)$
    \State Generate the measurement $\Vecy = \VecA\!\!\left(\!1{-}\sum_{k=1}^K
    \Vecu_k^{(0)}\!\right)\!\Vecf $
    \State Sample $\hat{\Vecy} \sim q_{\phi}(\Vecy|\hat{\Vecu}^{(0)})$
    \State Sample $t \sim \text{Uniform}(\{1, \ldots, T\})$
    \State Sample $\epsilon \sim \mathcal{N}(0, I)$
    \State Compute $\nabla \left[ \sum_{k=1}^{K} \left\| \epsilon - \epsilon_{\theta} \left( \sqrt{\bar{\alpha}_t} \Vecu_k^{(0)} + \sqrt{1 - \bar{\alpha}_t} \epsilon, t, \hat{\Vecy} \right) \right\|^2  \right]$; 
    \State Perform gradient descent.
\Until{converged}
\end{algorithmic}
\label{alg:training_ssd}
\end{algorithm}

\subsection{Inference Algorithm}
Given $\Vecy$ inference is performed using SSD according to \Cref{alg:inference_ssd}.

\begin{algorithm}[h]
\caption{SSD Inference algorithm}
\begin{algorithmic}[1]
    \State Given the soft shadow photograph $\Vecy$
    \State Encode the photograph: $\hat{\Vecy} \gets F_{\alpha}(\Vecy)$ 
    \State Sample noise from Gaussian distribution $ \{\Vecu_k^{(T)}\}_{k=1}^K \gets \hat{\Vecu}^{(T)}  \sim \mathcal{N}(0, I)$
    \For{$t = T, \ldots, 1$} 
        \State Denoise 
        \State Sample $\{\Vecu_k^{(t-1)}\}_{k=1}^K \sim p_{\theta}(\{\Vecu_k^{(t-1)}\}_{k=1}^K|\{\Vecu_k^{(t)}\}_{k=1}^K, \hat{\Vecy})$
    \EndFor
    \State \Return $\{\Vecu_k^{(0)}\}_{k=1}^K$
\end{algorithmic}
\label{alg:inference_ssd}
\end{algorithm}

\subsection{Implementation Details}

\paragraph{\textbf{Soft Shadow Image Encoder.}} The architecture of our encoder follows that of crossformer \cite{crossformer} with a number of transformer blocks (2, 4, 8, 2), shown in \Cref{tab:Crossformer}.

\paragraph{\textbf{Dataset Preparation.}} We simulated
262,000 measurements and sampled the corresponding pointclouds. From them, we randomly selected 3,000 examples as the test set, the remaining 259,000 are used as the training set.

\paragraph{\textbf{Diffusion Process.}} The number of steps in the diffusion process is $T = 256$. We set the variance schedules $\beta_t$ following a cosine schedule.

The training parameters are shown in \Cref{tab:training_paramters}.

\begin{figure}[h]
    \centering
    \includegraphics[width=0.5\textwidth]{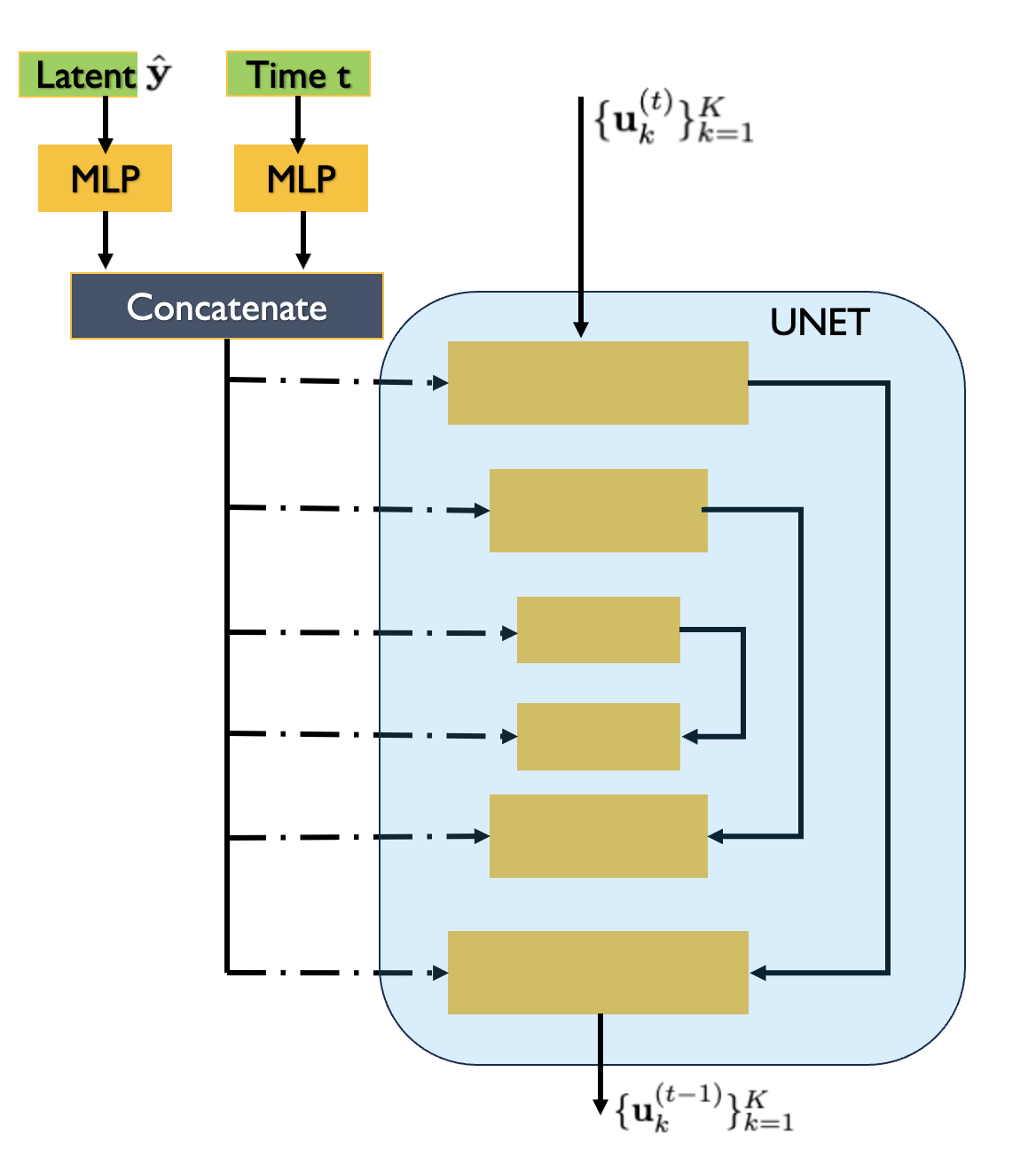}
    \caption{SSD UNet Module}
    \label{fig:SSD UNet Module}
\end{figure}

\paragraph{\textbf{U-Net-Based Diffusion Model.}} The model architecture is based on the Unet design, as detailed in \Cref{fig:SSD UNet Module}. This architecture is particularly suited for handling point cloud data due to its ability to efficiently process spatial hierarchies and feature representations. In our implementation, the model is trained on point cloud tensors with a shape of (2048, 3), representing a set of 2048 points in a three-dimensional space.

To enhance the model's capability in reconstructing detailed point clouds, we employ a sampling strategy that generates a denser point cloud. Specifically, during the reconstruction phase, we sample a higher number of points, aiming for 15,000 points. This approach allows for generating a more detailed and dense representation of the reconstructed occluding structure, capturing finer nuances and structures that are crucial for high-fidelity mesh representation.

\begin{table}[h]
\centering
\caption{ CrossFormer Model for Soft Shadow Image Encoding (CrossFormer-T)}
\begin{tabular}{lccc}
\toprule
Stages & Output Size & Layer Name & CrossFormer-T \\
\midrule
& & Cross Embed. & Kernel size: $\begin{bmatrix}
4 \times 4 \\
8 \times 8 \\
16 \times 16 \\
32 \times 32 \\
Stride=4 \\
\end{bmatrix}$\\
Stage-1 & 32 × 32 & SDA/LDA/MLP & 
$ \quad \quad \quad \quad \quad \quad \begin{bmatrix}
D1 = 64 \\
H1 = 2 \\
G1 = 7 \\
I1 = 8 \\
\end{bmatrix} \times 2$ \\
\midrule
&&Cross Embed. & Kernel size: 4 × 4, Stride=4 (S1 = 56) \\
Stage-2 & 16 × 16 & SDA/LDA/MLP & 
$\quad \quad \quad \quad \quad \quad  \begin{bmatrix}
D2 = 128 \\
H2 = 4 \\
G2 = 7 \\
I2 = 4 \\
\end{bmatrix} \times 4$ \\
\midrule
&&Cross Embed. & Kernel size: 2 × 2, Stride=2 (S2 = 28) \\
Stage-3 & 8 × 8 & SDA/LDA/MLP & 
$\quad \quad \quad \quad \quad \quad\begin{bmatrix}
D3 = 256 \\
H3 = 8 \\
G3 = 7 \\
I3 = 2 \\
\end{bmatrix} \times 8$ \\
\midrule
& & Cross Embed. & Kernel size: 2 × 2, Stride=2 (S3 = 14) \\

Stage-4 & 4 × 4 & SDA/LDA/MLP &
$\quad \quad \quad \quad \quad \quad\begin{bmatrix}
D4 = 512 \\
H4 = 16 \\
G4 = 7 \\
I4 = 1 \\
\end{bmatrix} \times 2$ \\
\midrule
&&Cross Embed. & Kernel size: 2 × 2, Stride=2 (S4 = 7) \\

\midrule
Head & 1 × 1 & Avg Pooling Kernel size: 4 × 4\\
\midrule
&&Linear Latent: 512 \\
\bottomrule
\label{tab:Crossformer}
\end{tabular}
\end{table}

\begin{table}[h]
\centering
\caption{SSD Training Details}
\begin{tabular}{@{}ll@{}}
\toprule
Parameter & Value \\
\midrule
Base channels & 128 \\
Optimizer & Adam\\
Channel multipliers & 1, 2, 4,8 \\
Learning rate & \(1 \times 10^{-4}\) \\
Blocks per resolution & 2 \\
Batch size & 192 \\
Attention resolutions & 256, 128 \\
EMA & 0.9999 \\
Attention Number of heads & 8 \\
Dropout & 0.0 \\
Conditioning embedding dimension & 512 \\
Training hardware & 3 × A100(80G) \\
Conditioning embedding MLP layers & 1 \\
Time embedding MLP layers & 1 \\
Training Iterations & 600000 \\
Diffusion noise schedule & Cosine \\
Weight decay & 0.01 \\
Sampling timesteps & 256 \\
Loss & Mean Squared Error (L2) \\
\bottomrule
\label{tab:training_paramters}
\end{tabular}
\end{table}

\clearpage

\section{Results}
\label{sup_sec:expt}
\subsection{Background Lights and Distortions}
\begin{figure}[ht]
\centering
    \includegraphics[width=0.8\linewidth]{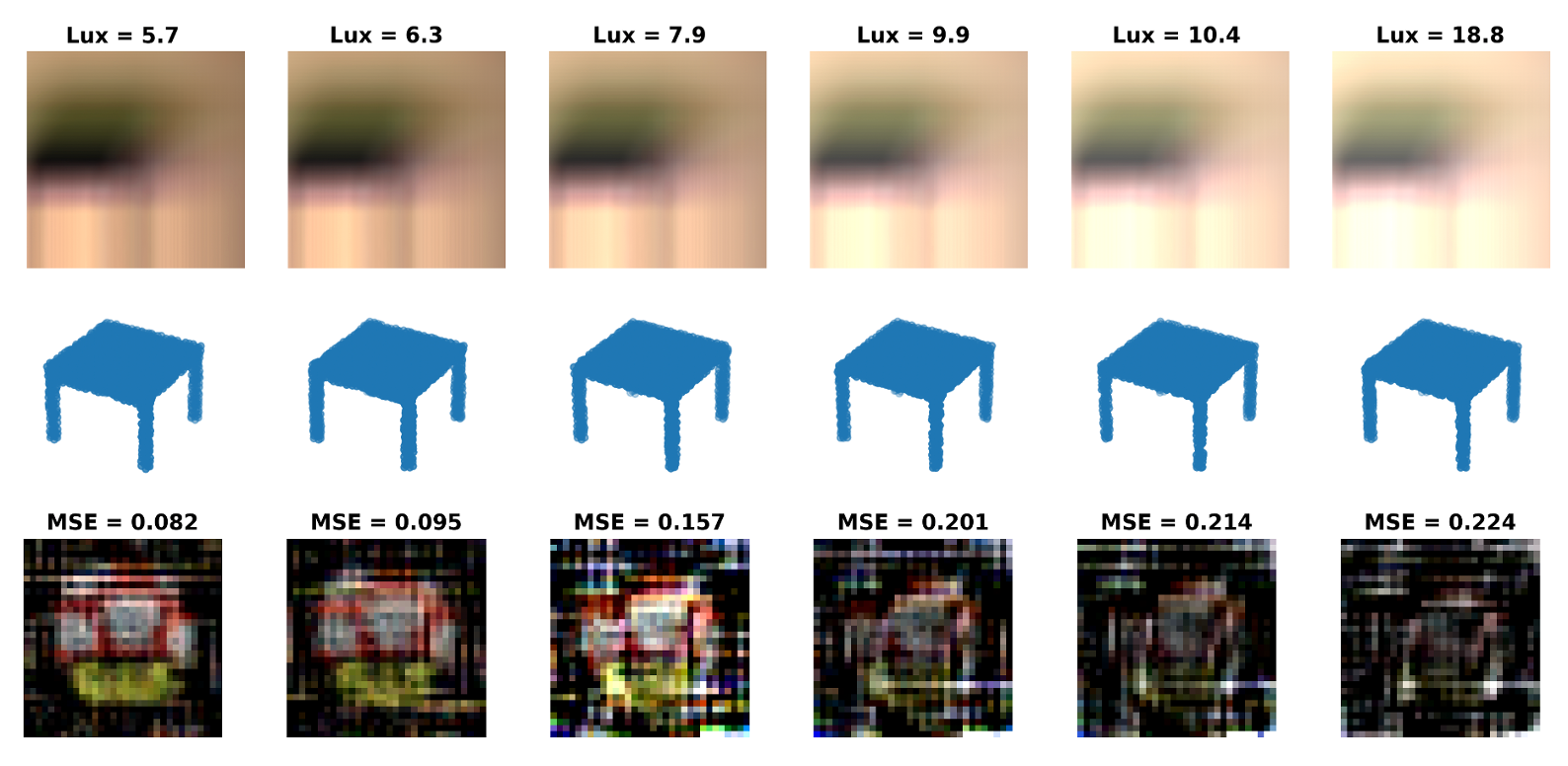}
    \caption{Experimental Results with Varying Background Lights}
    \label{fig:ssdrealbackgroundresults}
\end{figure}

\begin{figure}[ht]
\centering
    \includegraphics[width=0.8\linewidth]{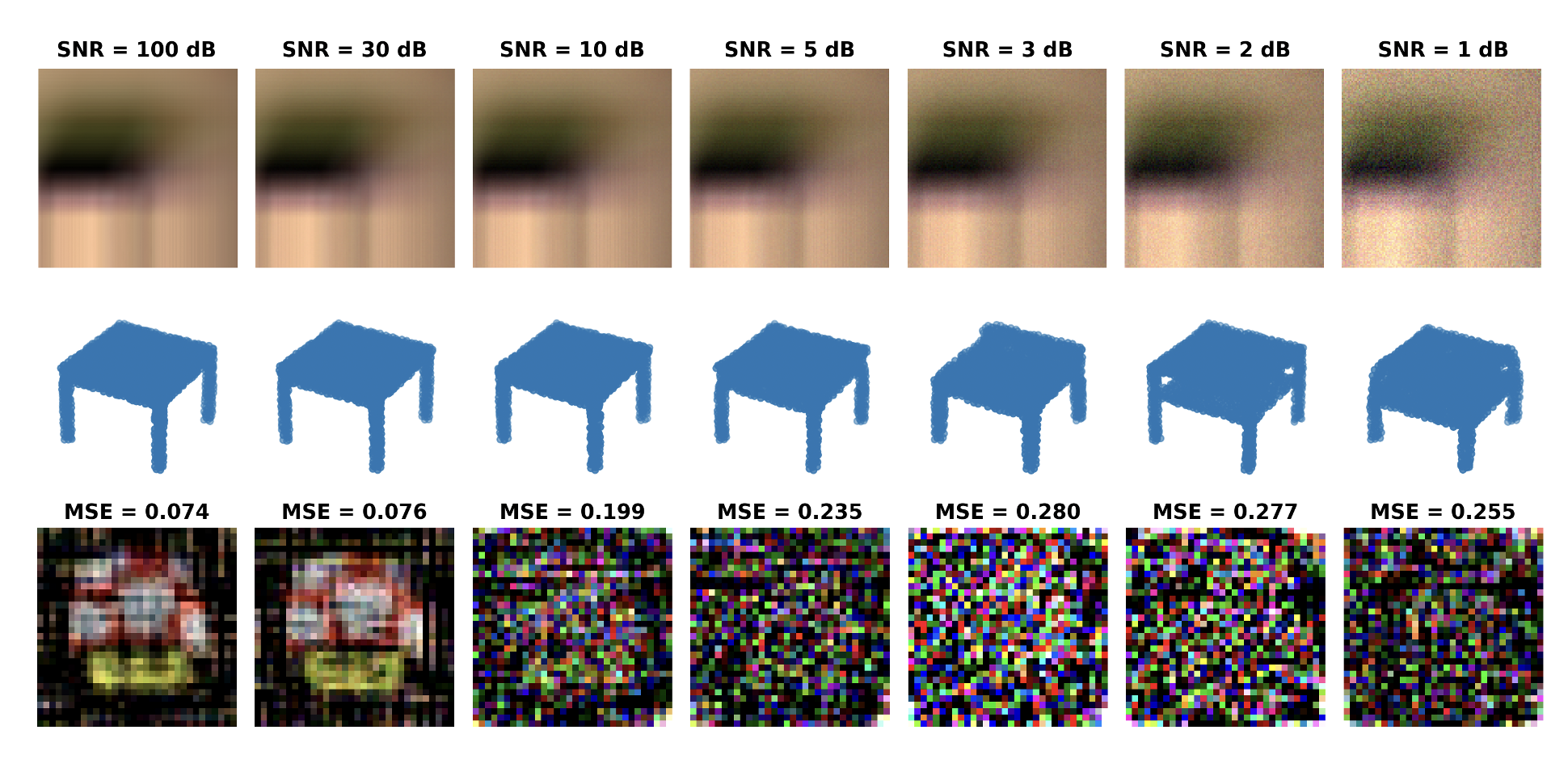}
    \caption{Experimental Results with Varying Signal-to-Noise Ratio (SNR)}
    \label{fig:ssdrealsnrresults}
\end{figure}

\begin{figure}[ht]
\centering
    \includegraphics[width=1\linewidth]{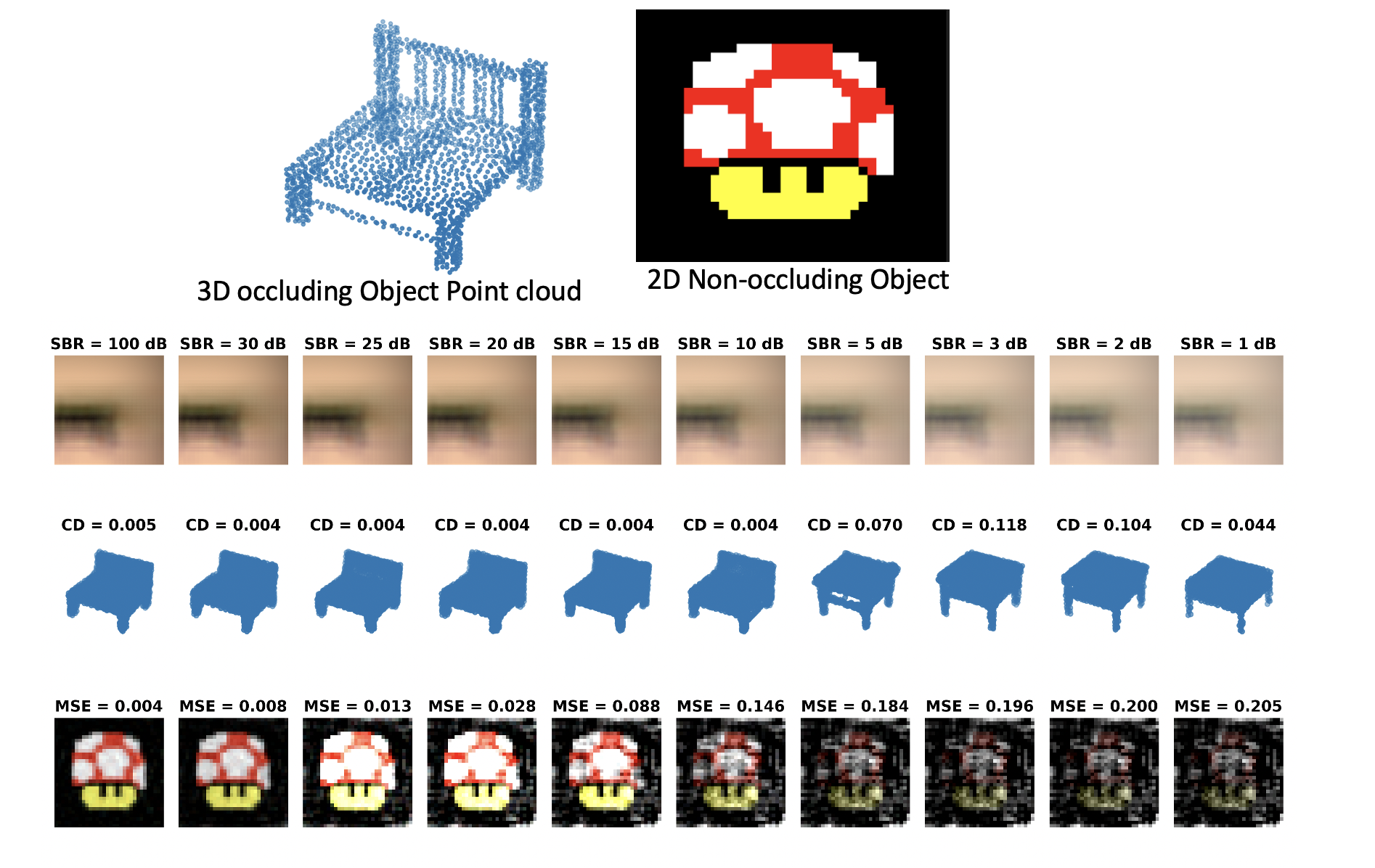}
    \caption{Simulated Results with Varying Signal-to-Background Ratio (SBR)}
    \label{fig:ssdsimulatedbackgroundresults}
\end{figure}

\begin{figure}[h]
\centering
    \includegraphics[width=1\linewidth]{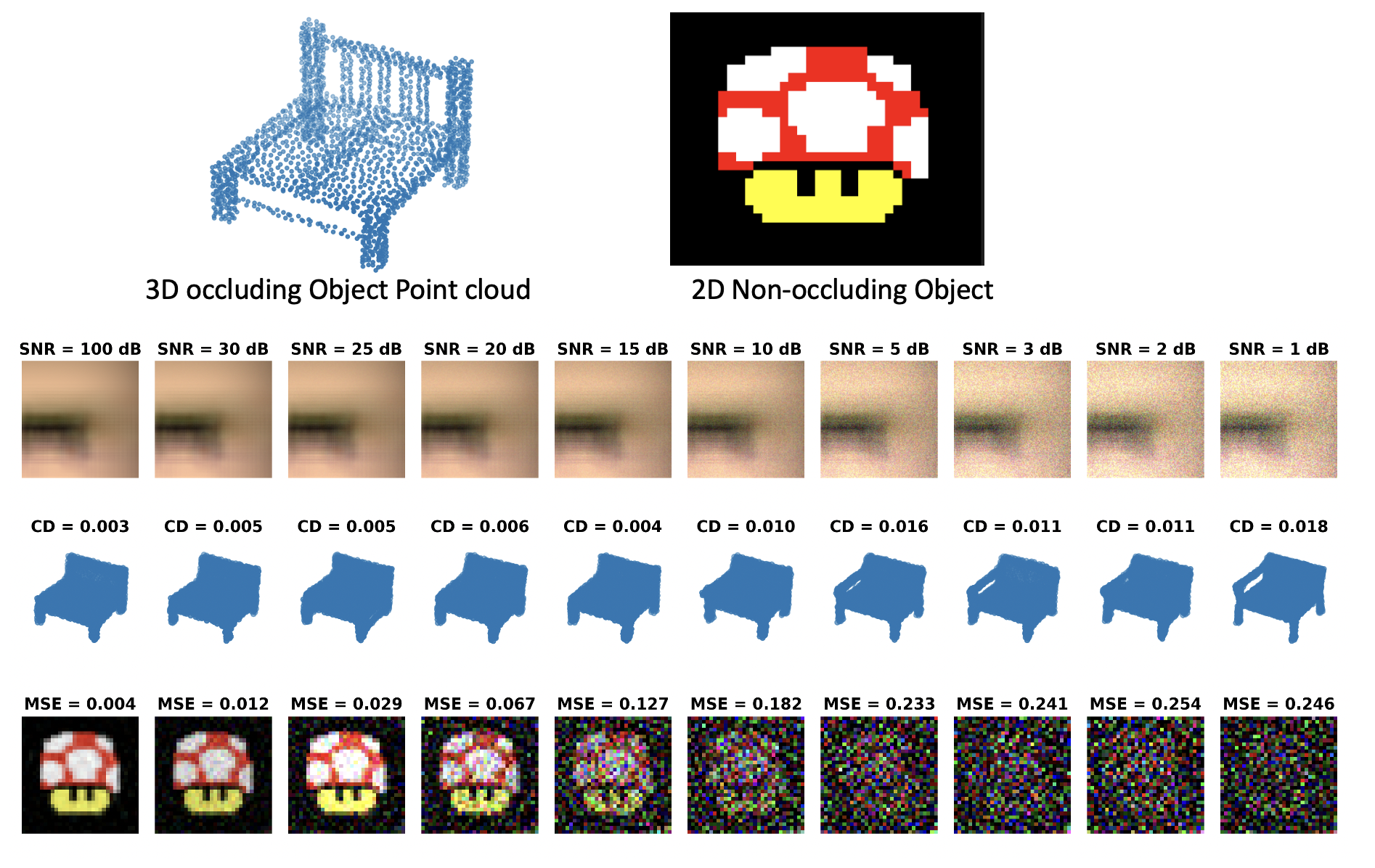}
    \caption{Simulated Results with Varying Signal-to-Noise Ratio (SNR)}
    \label{fig:ssdsimulatedsnrresults}
\end{figure}

To investigate the influence of background and noise on our method, we incorporated a controlled light source within the experimental setup. The quantitative results are presented in Figure 6(b) of the main paper. Figure \ref{fig:ssdrealbackgroundresults} presents the qualitative results of this experiment. We also simulated a noise process on the real measurement and the result is presented in Figure \ref{fig:ssdrealsnrresults}. Simulated examples are shown in Figures \ref{fig:ssdsimulatedbackgroundresults} and \ref{fig:ssdsimulatedsnrresults}

\clearpage
\subsection{Shape Reconstructions}
\begin{figure}[h]
\centering
    \includegraphics[width=1\linewidth]{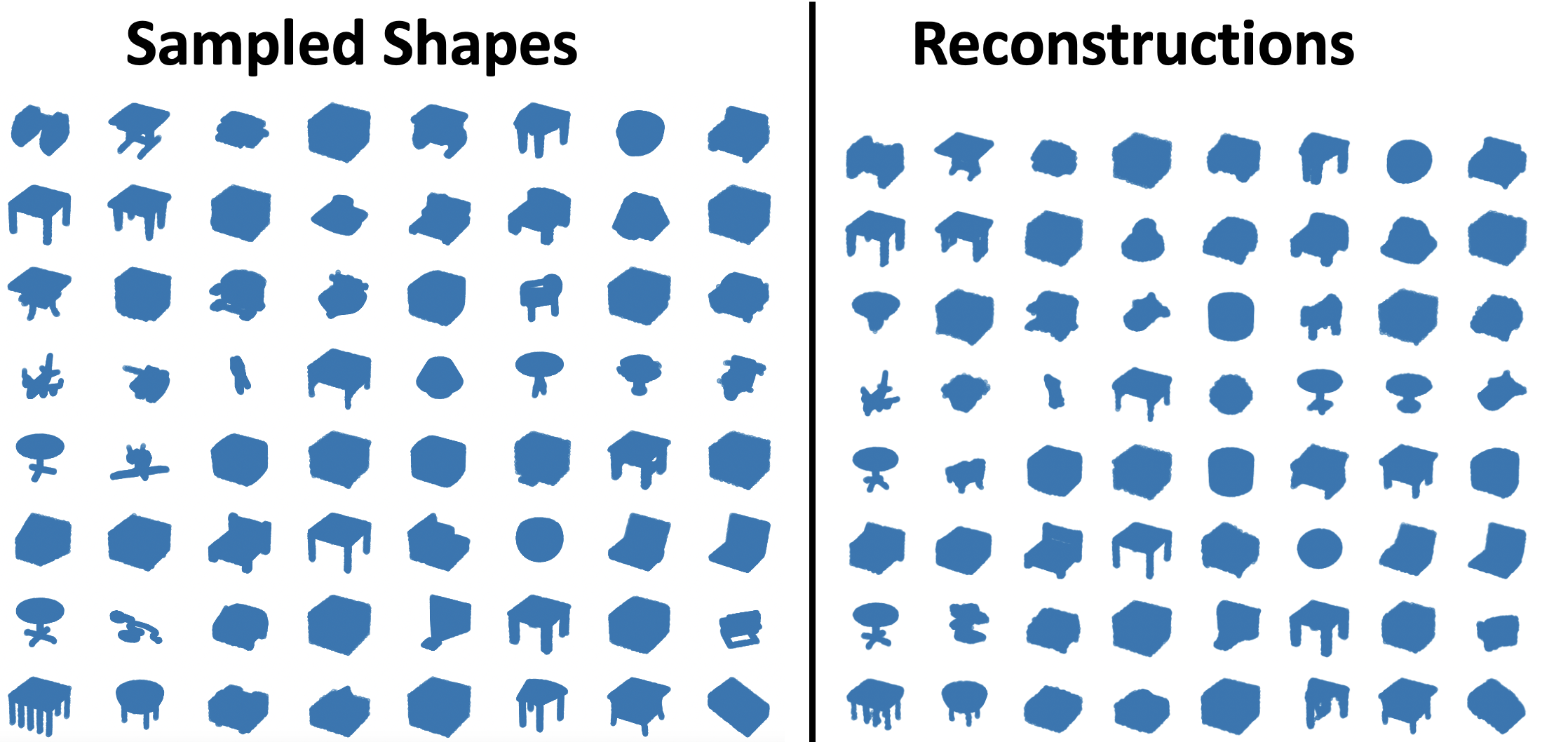}
    \caption{Reconstructed shapes outside the Training dataset. }
    \label{fig:samplesshapes}
\end{figure}

Additional results are provided here to demonstrate the performance of the proposed SSD model on a variety of shape classes. We sampled 64 meshes outside our training dataset, and simulated the penumbra photograph using random 2D non-occluding objects. The result of this experiment is shown in Figure \ref{fig:samplesshapes}.

\begin{figure}[h]
\centering
    \includegraphics[width=0.8\linewidth]{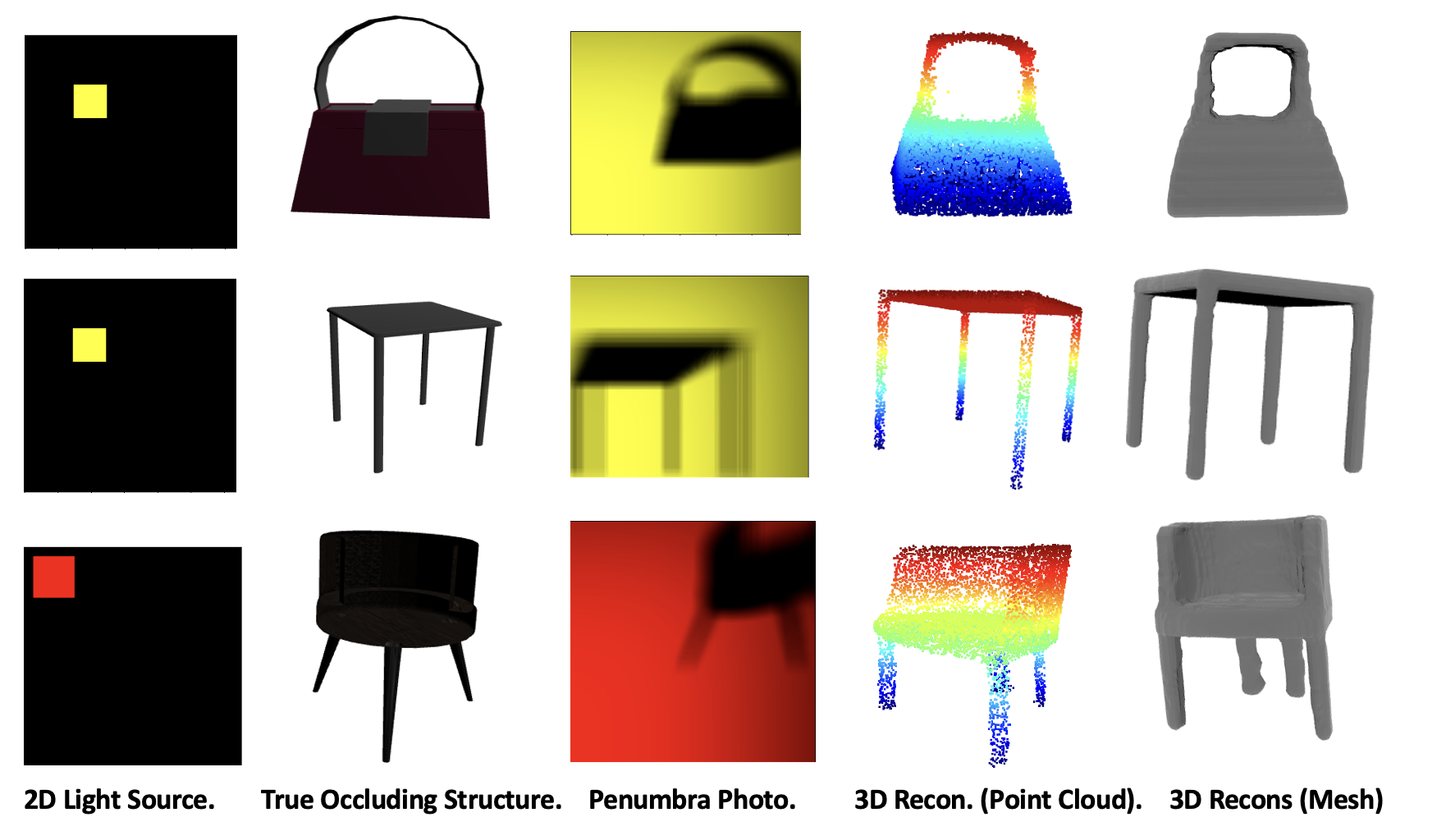}
    \caption{SSD Reconstruction from Hard Shadows.}
    \label{fig:ssdresults4}
\end{figure}

\begin{figure}[h]
\centering
    \includegraphics[width=1\linewidth]{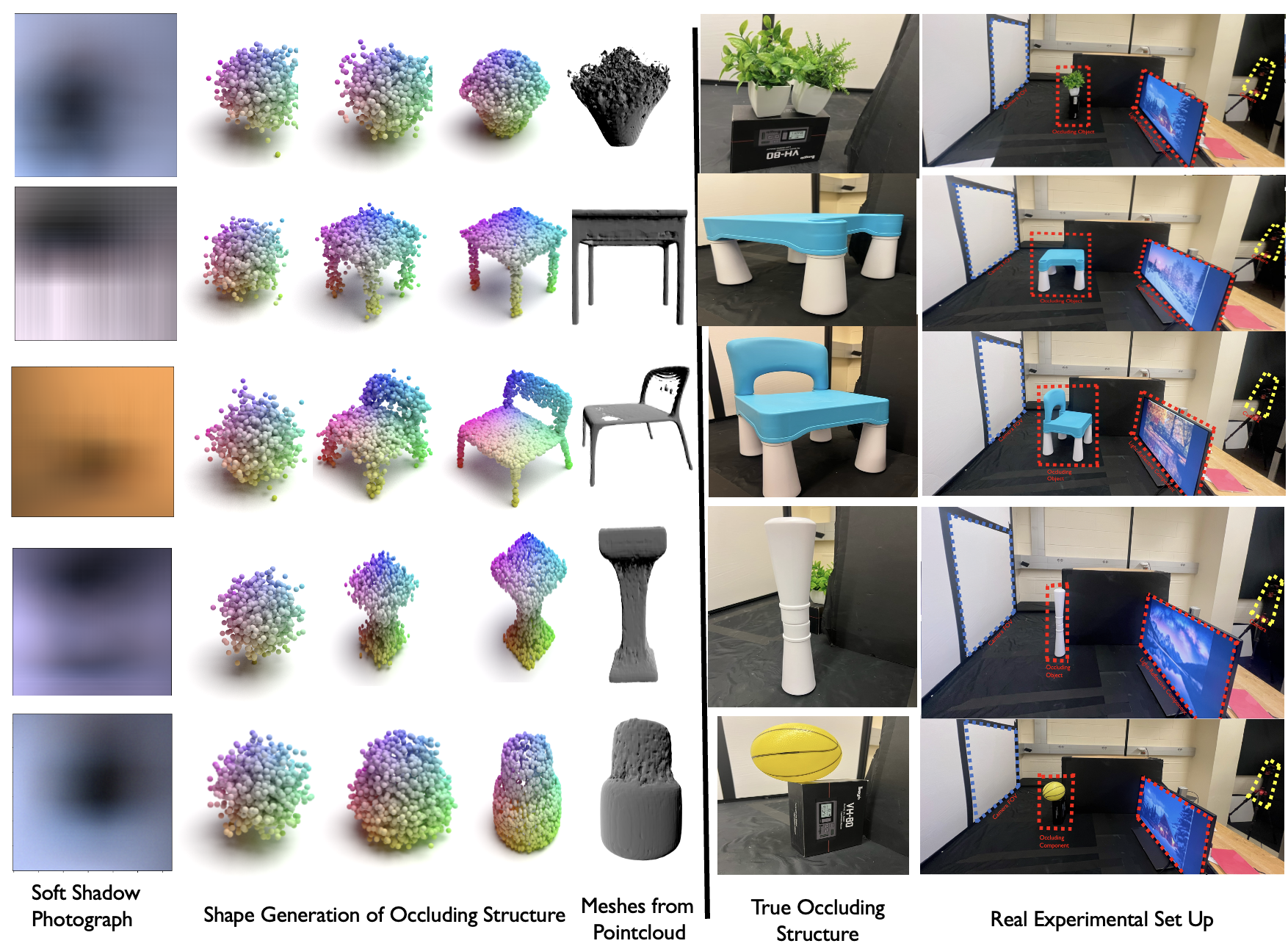}
    \caption{Results from SSD in Real Experimental Setup.}
    \label{fig:real_ssd}
\end{figure}

\begin{figure}[h]
\centering
    \includegraphics[width=1\linewidth]{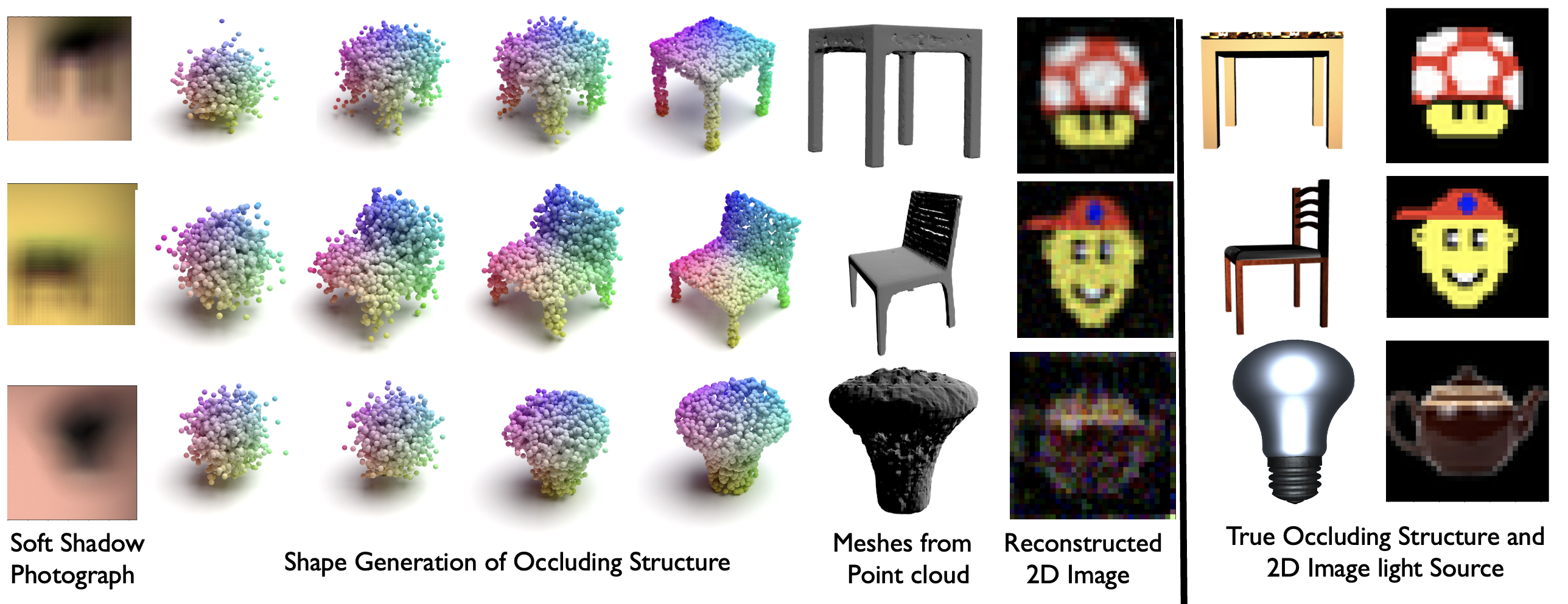}
    \caption{Results from SSD in Simulation.}
    \label{fig:ssdresults}
\end{figure}

\begin{figure}[h]
\centering
    \includegraphics[width=1\linewidth]{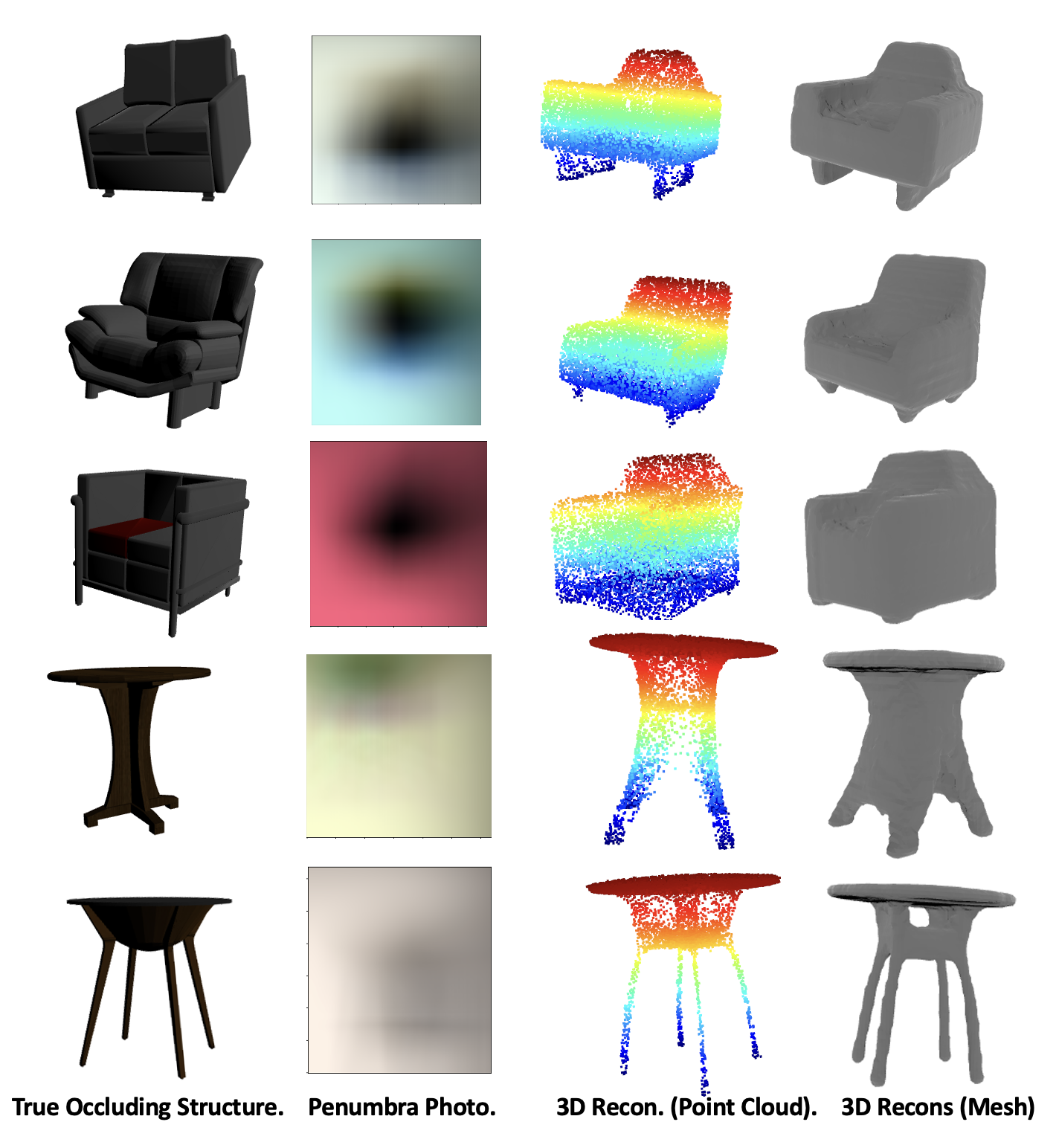}
    \caption{Results from SSD in Simulation.}
    \label{fig:ssdresults2}
\end{figure}

\begin{figure}[h]
\centering
    \includegraphics[width=1\linewidth]{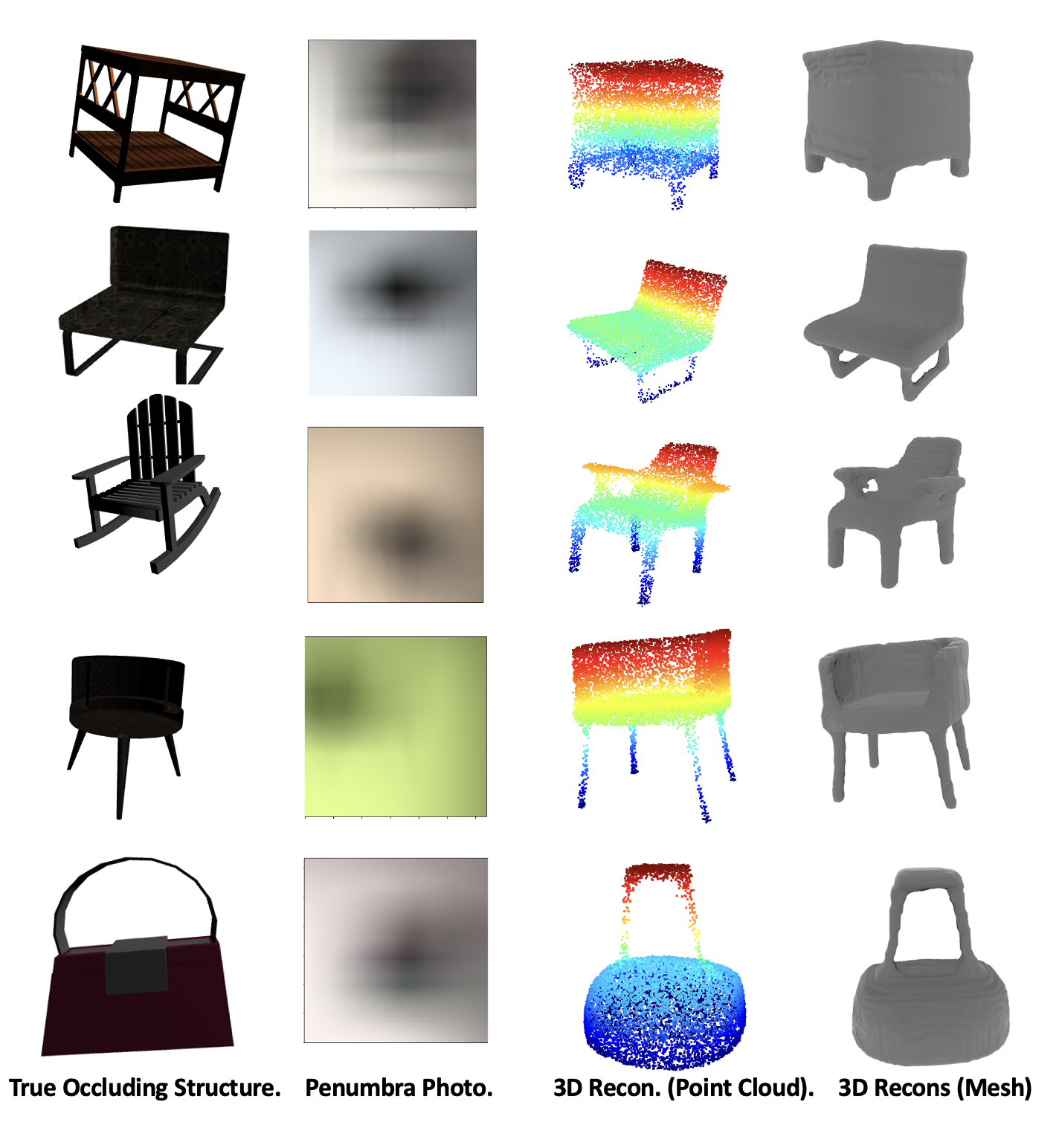}
    \caption{Results from SSD in Simulation.}
    \label{fig:ssdresults3}
\end{figure}

We recall that the model is trained in simulation. Results for simulated experimental data are shown in \Cref{fig:ssdresults,fig:ssdresults2,fig:ssdresults3,fig:ssdresults4}, while additional results based on real measured data are shown in \Cref{fig:real_ssd}.
\vspace{-8mm}

\clearpage

\end{document}